\documentclass[article]{elsarticle}

\usepackage{hyperref}
\usepackage[utf8]{inputenc}
\usepackage{amsmath}
\usepackage{version}
\usepackage{floatrow}
\usepackage{float}
\usepackage{multirow}
\usepackage{caption}
\usepackage{array}
\usepackage{booktabs}
\usepackage{graphicx}
\usepackage{url}
\usepackage[tight, footnotesize]{subfigure}

\newcommand{\xii}{\mathbf{x_i}}
\newcommand{\xjj}{\mathbf{x_j}}

\newcommand{\xjt}{\mathbf{x_{j,t}}}
\newcommand{\LL}{\mathcal{L}}
\newcommand{\KK}{\mathcal{K}}
\newcommand{\PP}{\mathcal{P}}
\newcommand{\FF}{\mathcal{F}}
\newcommand{\SSS}{\mathcal{M}}

\graphicspath{{./img/}}

\begin{document}

\begin{frontmatter}

\title {An Optimized PatchMatch for Multi-scale and Multi-feature Label Fusion}


\author[add1,add2,add3,add4,add5]{Rémi Giraud  \corref{cor1}}
\author[add1,add2,add5]{Vinh-Thong Ta}
\author[add3,add4]{Nicolas Papadakis}
\author[add6]{José V. Manjón}
\author[add7]{D. Louis Collins}
\author[add1,add2]{Pierrick Coupé}
\author[]{and the Alzheimer's Disease Neuroimaging Initiative
\protect\footnote{Data used in preparation of this article were obtained from the Alzheimer's
Disease Neuroimaging Initiative (ADNI) database (adni.loni.usc.edu). As such, the
investigators within the ADNI contributed to the design and implementation of
ADNI and/or provided data but did not participate in analysis or writing of this
report. A complete listing of ADNI investigators can be found at: 
\url{http://adni.loni.use.edu/wp-content/uploads/how\_to\_apply/ADNI\_Acknowledgement\_List.pdf.}}}
\address[add1]{Univ. Bordeaux, LaBRI, UMR 5800, PICTURA, F-33400 Talence, France.\\}
\address[add2]{CNRS, LaBRI, UMR 5800, PICTURA, F-33400 Talence, France.\\}
\address[add3]{Univ. Bordeaux, IMB, UMR 5251, F-33400 Talence, France.\\}
\address[add4]{CNRS, IMB, UMR 5251, F-33400 Talence, France.\\}
\address[add5]{Bordeaux INP, LaBRI, UMR 5800, PICTURA, F-33600 Pessac, France.\\}
\address[add6]{Instituto de Aplicaciones de las Tecnologías de la Información y de las Comunicaciones
Avanzadas (ITACA), Universitat Politècnica de València, Camino de Vera s/n, 46022 Valencia,
Spain.\\}
\address[add7]{McConnell Brain Imaging Centre, Montreal Neurological Institute,
McGill University, Montreal, Canada.\\}

\begin{abstract}
Automatic segmentation methods are important tools for quantitative
analysis of Magnetic Resonance Images (MRI). 
Recently, patch-based label fusion approaches have demonstrated
state-of-the-art segmentation 
accuracy. 
In this paper, we introduce a new patch-based label fusion framework to perform segmentation of anatomical structures. 
The proposed approach uses an Optimized PAtchMatch Label fusion (OPAL) strategy that drastically reduces the computation time required for the search of similar patches.
The reduced computation time of OPAL  opens the way for new strategies and facilitates processing on large databases. 
In this paper, we investigate new perspectives offered by OPAL, by introducing a new multi-scale and multi-feature framework. 
During our validation on hippocampus segmentation we use two datasets: young adults in 
the ICBM cohort and elderly adults in the EADC-ADNI dataset. 
For both, OPAL is compared to state-of-the-art methods. 
Results show that OPAL obtained the highest median Dice
coefficient ($89.9\%$ for ICBM and $90.1\%$ for EADC-ADNI). 
Moreover, in both cases, 
OPAL produced a segmentation accuracy similar to inter-expert variability. 
On the EADC-ADNI dataset, we compare the hippocampal volumes obtained by manual and automatic segmentation. 
The volumes appear to be highly correlated that enables to perform more accurate separation of pathological populations.

\end{abstract}


\end{frontmatter}


\section{Introduction}

Magnetic Resonance Imaging (MRI) has become an essential tool in medical analysis,
especially in the study of the human brain.
The segmentation of MRI brain structures is a necessary step for many clinical 
applications. The manual segmentation of structures in MRI by clinical experts
is still considered as the gold standard. 
However,  manual labeling is a highly tedious and very time consuming task. 
Moreover, the manually generated segmentations are subject 
to inter- and intra-rater variability. Therefore, designing fast, accurate and reliable automatic segmentation methods is a challenging work in quantitative MRI analysis.

In the past decade, 
several paradigms were proposed to automatically perform brain segmentation.
First, atlas-based methods involving nonlinear
registration of a labeled atlas to the subject were
proposed ~\cite{collins1995automatic, babalola2009}.
Once the atlas is matched to the subject image, the segmentation is achieved by warping the atlas labels to the target
image space. Such atlas-based methods have been widely used due to their robustness and the ease of integration of expert priors. 
However, atlas-based methods may not sufficiently capture inter-subject variability due to the one-to-one mapping assumption between the atlas and the subject anatomy.
Consequently, atlas-based methods are subject to registration errors since in general such mapping does not exist.  

In order to minimize registration errors, template warping techniques based 
on a training library of manually labeled templates were introduced.
The simplest method based on a library of training templates is the best-template approach \cite{barnes2008comparison}.
The main idea is to reduce the anatomical distance between a selected template and the subject to be segmented in order to improve registration accuracy. 
First, the most similar template is selected in the training library. Then, this template is nonlinearly registered to the subject. 
Finally, the estimated nonlinear transformation is applied to the manually segmented labels in the selected template to obtain the final segmentation. 
While the selection of the most similar template compared to an \textit{a priori} 
fixed atlas may improve segmentation results, the best template strategy is still subject 
to registration errors and leads to sub-optimal results.

A significant improvement has been obtained with the introduction of multi-template approaches.
Such methods merge information from 
several similar training templates instead of using a single template to achieve better
segmentation. In such methods, the registration errors resulting from inter-subject variability are considered as a random variable,
thus reducing segmentation error by using several atlases \cite{rohlfing2004, heckemann2006automatic}.
Since its introduction, many approaches have been proposed to improve the label fusion step, such as preselection of 
most similar template following by majority voting \cite{aljabar2009, collins2010towards, cardoso2013},
 intensity models 
\cite{wolz2009, lotjonen2010fast}, fusion techniques
 with local weighted label fusion \cite{artaechevarria2009, khan2011,sabuncu2010generative} or 
 systematic bias correction using a learning-based method \cite{wang2011}. Multi-templates matching approaches 
demonstrated competitive 
segmentation accuracy at the expense of an important computational burden 
resulting from multiple nonlinear registrations, \emph{i.e.}, up to
several hours. 


Recently, a nonlocal
patch-based label fusion (PBL) method \cite{coupe2011patch} has been
proposed for reducing the computational burden of multi-templates based methods.
Instead of performing multiple nonlinear registrations, the PBL method relies on 
the comparison of patches (centered neighborhood around a voxel) 
which only requires an affine alignment of the subject and the training templates.
The patch comparisons performed between the current image patch and training patches,
are used to assign a weight to the manual labels according to patch similarity.
The search for similar training patches is based on a nonlocal strategy in order to
better capture registration inaccuracies and 
to efficiently handle the inter-subject variability. 
PBL overcomes the one-to-one mapping assumption of multi-template warping methods thanks to a well-defined one-to-many mapping model. 
Finally, the PBL approach produces state-of-the-art segmentation
accuracy with limited computation time,
\emph{i.e.}, several minutes. 

Since its introduction, 
the PBL approach has been intensively
studied and many improvements have been proposed.
First, PBL can be combined with other methods such as multi-template warping  \cite{rousseau2011supervised}, 
active appearance models \cite{hu2014nonlocal} or level sets \cite{wang2014segmentation}. 
Moreover, other improvements have been proposed using multi-resolution framework \cite{eskildsen2012beast}, 
discriminative dictionary learning and sparse coding
\cite{tong2013segmentation}, or generative probability models
\cite{wu2013generative}. 
However,
PBL still suffers from several limitations. 
First, the search for similar 
patches is still computationally expensive. 
Although preselection of templates and patches
\cite{coupe2011patch} or
multi-scale strategies \cite{eskildsen2012beast}  
have been proposed, an important amount of computation remains
dedicated to the search for
similar patches in the training library.  
Secondly, the template preselection step
can prevent finding
the most similar patches existing in the library. 
By selecting  training templates according 
to a global similarity measure between the subject and the template,
the template preselection step
is likely to remove relevant parts of the training library, possibly 
leading to sub-optimal results. 
Finally, in PBL, patch comparisons are performed between the 
current patch and training patches. The relevance of the match
is then weighted depending on the
similarity between the two patches. 
However, weights are assigned  
to a large number of training patches including many dissimilar patches.
Beyond inefficient computations dedicated to estimate negligible weights,  
these  dissimilar patches can decrease the segmentation accuracy 
\cite{tong2013segmentation}.   
Sparsity-based methods
tend to limit this issue but suffer from an important computational burden
\cite{tong2013segmentation, wu2013generative}.

In this paper, we first introduce a new Optimized PAtchMatch for Label fusion (OPAL) to address the limitations of 
previous PBL approaches in terms of computation time and search strategy of similar patches. 
The OPAL method is  able to find, in significantly less computations, similar patches over the entire training library without template or patch preselection. 
Originally, the PatchMatch (PM) \cite{barnes2009patchmatch} algorithm was introduced to efficiently 
find patch correspondences between two 2D images. For each patch within the first image,
an approximate nearest neighbor (ANN) is found within the second image.
The algorithm is based on a cooperative and randomized strategy resulting
in very low computation time, enabling near real-time processing.  
PM has been applied to medical imaging for super-resolution of cardiac
MRI \cite{shi2013cardiac}, but most PM applications concern 2D image editing problems. 
In this work, we investigate the use of PM for anatomical
structures segmentation using multi-templates training library. Thanks to our Optimized PM (OPM) algorithm, 
OPAL produces segmentations in 
a few seconds compared to previous PBL methods.  
Beyond computation time efficiency, OPAL complexity 
only depends on the size of the area to be processed within the subject. 
Consequently, our method does not require any preselection,
since the search of most similar patches is achieved over the entire
training library. Without training template or patch preselection,
similar patches can be found within the whole template library 
leading to higher segmentation accuracy.

The drastically reduced computation time of OPAL opens the way for new strategies and efficient processing of very large databases. 
In this paper, we investigate new perspectives offered by OPAL by introducing a new multi-scale and multi-feature framework. 
In our approach, several scales and features are analyzed at the same time before performing the label fusion.
First, the OPM is achieved with different patch sizes on each feature.
Then, we perform a late fusion of these independent estimators, 
each one providing different information on structure characteristics.
The description of the structures indeed depends on the considered patch size or the image features used.
By using multi-scale and multi-feature searches, the diversity of selected matches is improved which increases the segmentation accuracy.

The main contributions of this work are: 
(i) An adaptation of the PM algorithm to label fusion for anatomical
structure segmentation in 3D MRI,
including acceleration techniques such as constrained initialization,
parallel processing and optimized distance computation;
(ii) A novel late fusion strategy of multi-scale and multi-feature estimator maps;
(iii) An extensive OPAL validation on hippocampus segmentation on two datasets with
comparison to state-of-the-art methods in terms of
computation time and segmentation accuracy; 
and (iv) A comparison of the ability to separate populations,
based on hippocampal volumes obtained with manual and automatic segmentation.

\section{Methods}

\subsection{Fast Nearest Neighbor Matching}

In the PBL method, the first step consists in finding, for each patch of the subject to segment, 
relevant matches, \emph{i.e.}, approximate nearest neighbors (ANN), within the training template library.
The two main issues of this method are the relevance of the selected patches and
the computational burden dedicated to this search.
In this work, we propose a fast patch-based nearest neighbor matching algorithm to
find highly similar patches, thus
addressing the computational costs usually associated with classic PBL techniques.

\subsubsection{The PatchMatch Algorithm}

The original PM algorithm \cite{barnes2009patchmatch} is a
fast and efficient approach that computes patch
correspondences (matches) between two 2D images (\emph{e.g.} $A$ \& $B$). 
The key point of this method is that good matches can be propagated to 
the adjacent patches within an image. This propagation, combined 
with random matches, leads to a very fast convergence with limited computational burden.
The core of the algorithm is based on three steps: initialization, propagation,
and random search.  
The initialization consists in randomly associating each
patch of $A$ with a corresponding patch in $B$, 
in order to obtain an initial ANN field. 
The two following steps are then performed iteratively in order to improve
the ANN field.
The propagation step uses
the assumption that when a patch $p$ centered on $\mathbf{x_i} = (x,y)\in A$
matches well with a patch $q$
centered on $\mathbf{x_j}\in B$, then the adjacent patches of $p \in A$ should
match well with the adjacent patches of $q \in B$.
The iterative process follows a scan order (from left to right, top to bottom) 
on even iterations and is reversed on odd iterations. Therefore, only recently processed pixels are 
selected to propagate good matches to their neighbors. 
For example, on even iterations, for a patch located at $\mathbf{x_i} = (x,y)\in A$, 
only the neighboring patches centered on $(x-1,y)$ and $(x,y-1)$ are considered during the propagation step.
Let $\mathbf{x_j'}\in B$ be the match of the patch centered on position $(x-1,y)\in A$. The candidate
to improve $p$ correspondence is the patch centered on $\mathbf{x_j'}+(1,0)\in B$.

Next, the random search step consists of
a random sampling around the current ANN to escape from local minima. 
The candidates are randomly selected 
within an exponentially decreasing search window centered on $\xjj $.
The propagation of good matches within the iterative process combined 
with random search, provides a very 
fast convergence of the algorithm in practice.

\subsubsection{\label{subsubsection:OPM} Optimized PatchMatch Algorithm}
In contrast to \cite{barnes2009patchmatch} where two 2D images are
considered, OPAL finds the patch 
correspondences between a 3D image $S$ and a library of $n$ 3D templates
$T=\{T_1, \dotsc, T_n\}$.
One advantage of the PM algorithm is that its complexity
 only depends on the size of image $A$ to process and not on the 
size of the compared image $B$, \emph{i.e.}, $T$ in the OPAL case.
This important fact enables OPAL to consider the entire image library
$T$ without any template preselection step at constant complexity in time.
Moreover, for each patch in $S$, OPAL computes the best $k$-ANN
matches in $T$ and not only one match as done in \cite{barnes2009patchmatch}.

\begin{figure}[h!]
\centering
\newcommand{\siz}{0.27\textwidth}
\centerline{
  \subfigure[CI]
  {\label{subfig:init}\fbox{\includegraphics[width=\siz,height=140pt]{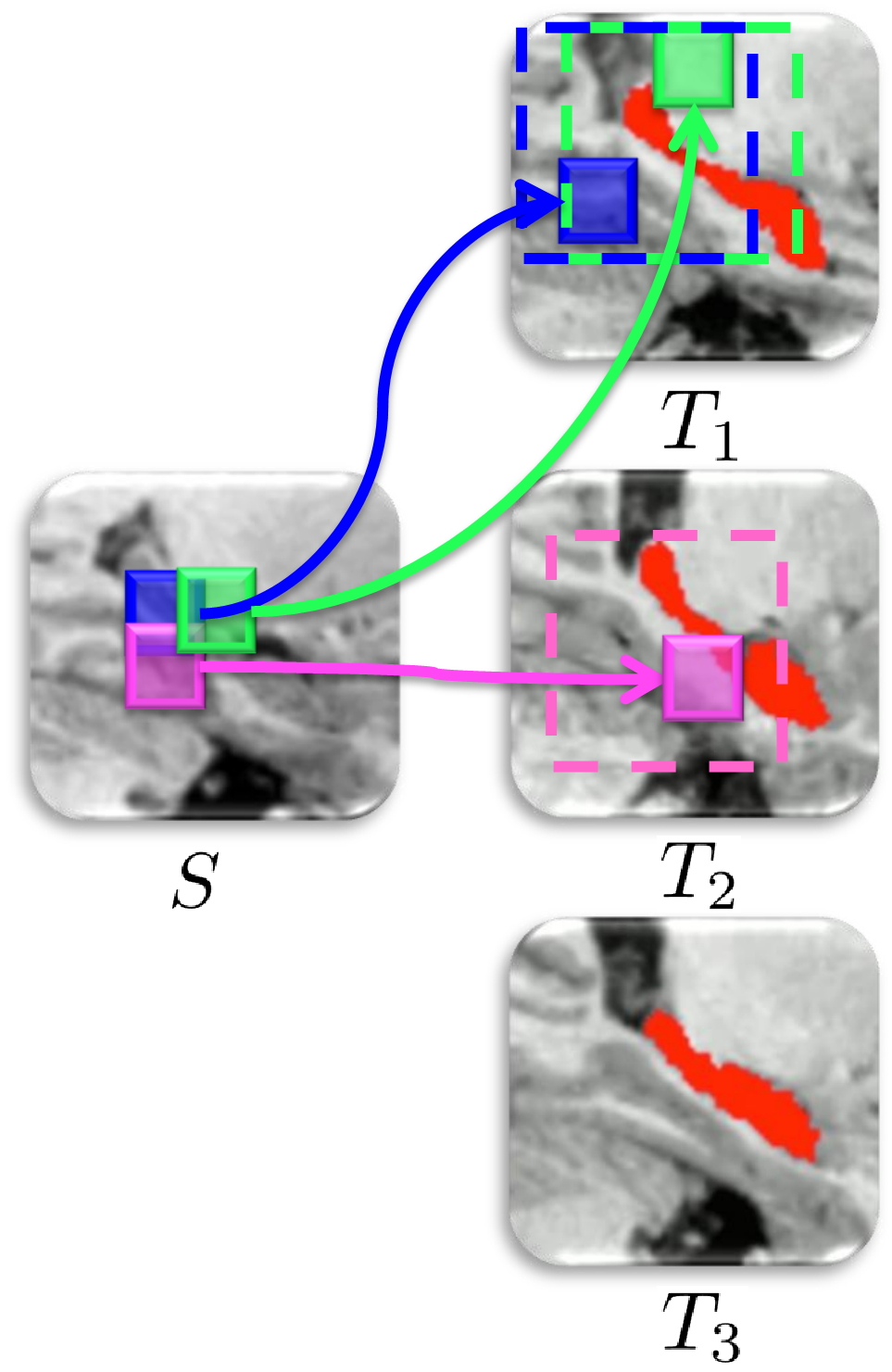}}}  
  \subfigure[PS for iteration $\#1$]
  {\label{subfig:prop:one}\fbox{\includegraphics[width=\siz,height=140pt]{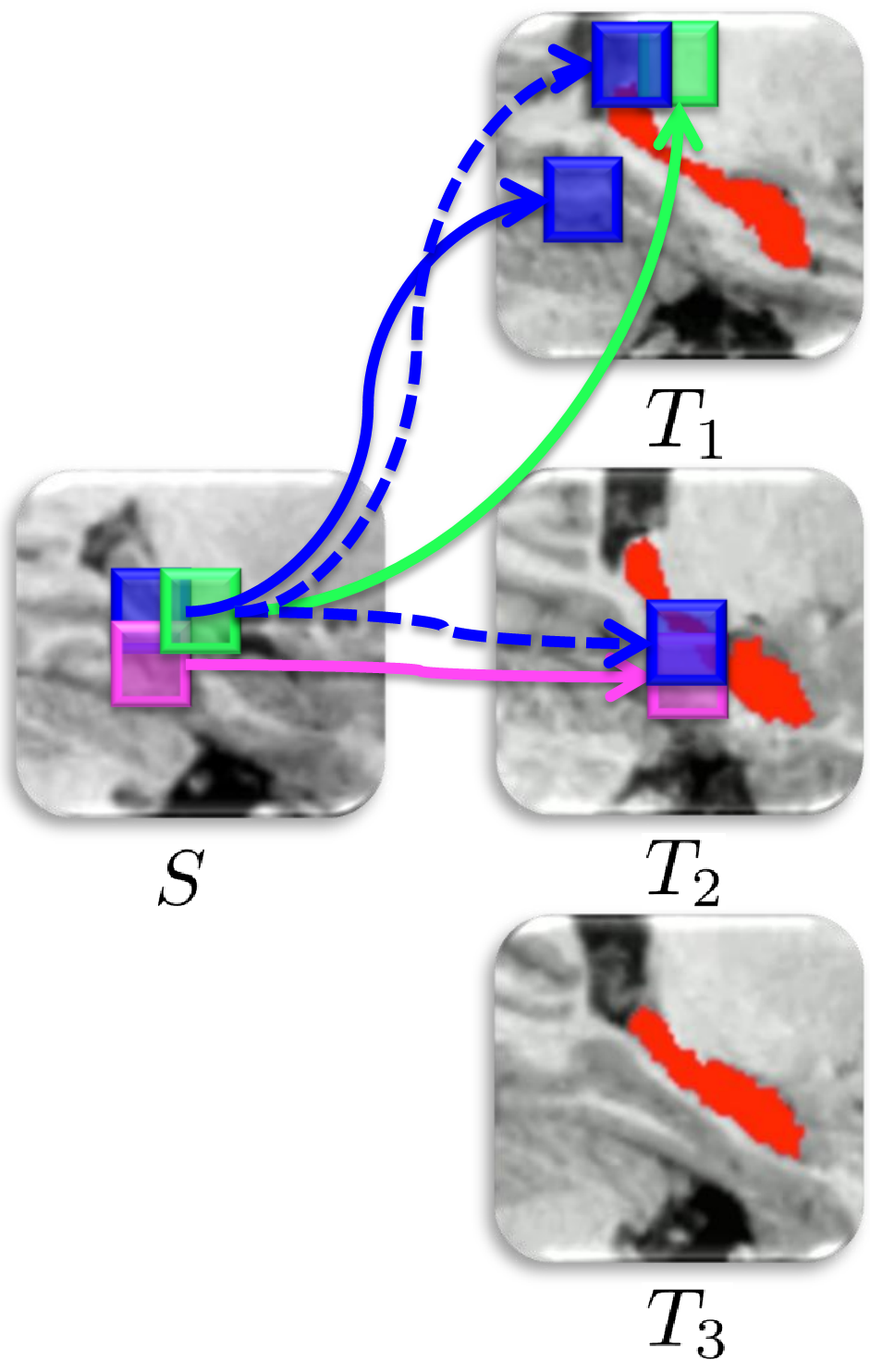}}}
  \subfigure[CRS for iteration $\#1$]
  {\label{subfig:rs:one}\fbox{\includegraphics[width=\siz,height=140pt]{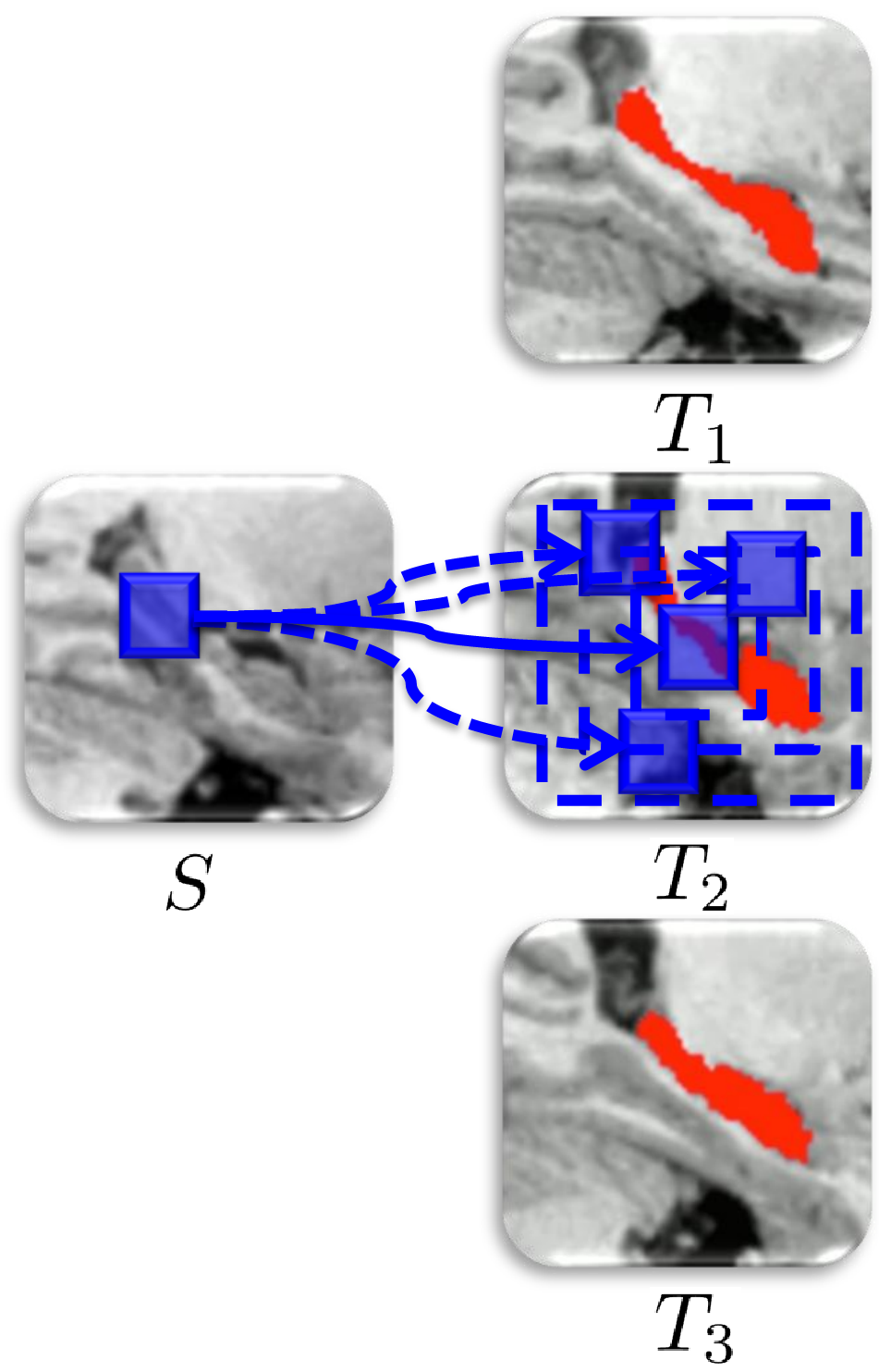}}}
}
\centerline{
  \subfigure[PS for iteration $\#2$]
  {\label{subfig:prop:two}\fbox{\includegraphics[width=\siz,height=140pt]{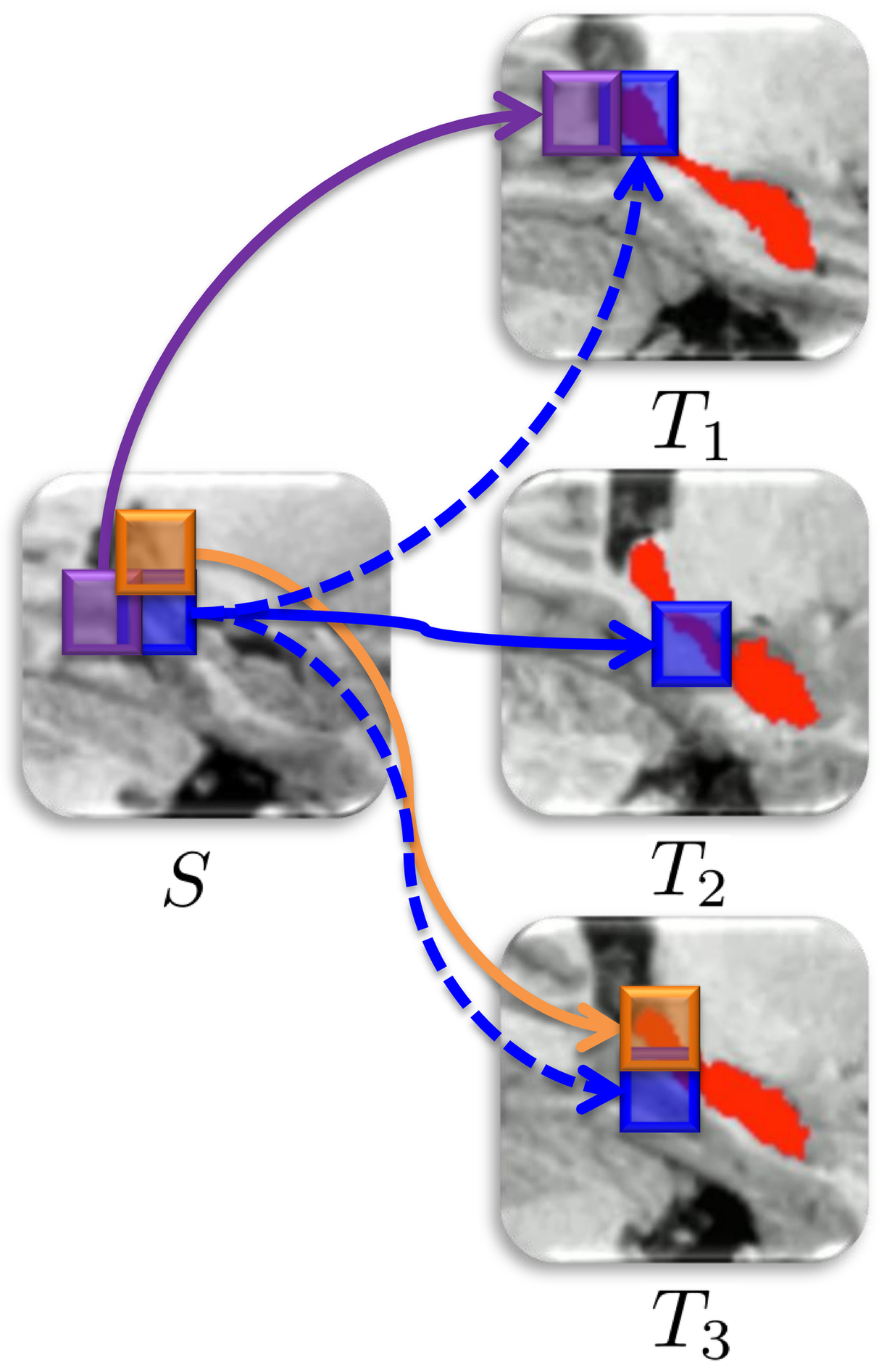}}}
  \subfigure[CRS for iteration $\#2$]
  {\label{subfig:rs:two}\fbox{\includegraphics[width=\siz,height=140pt]{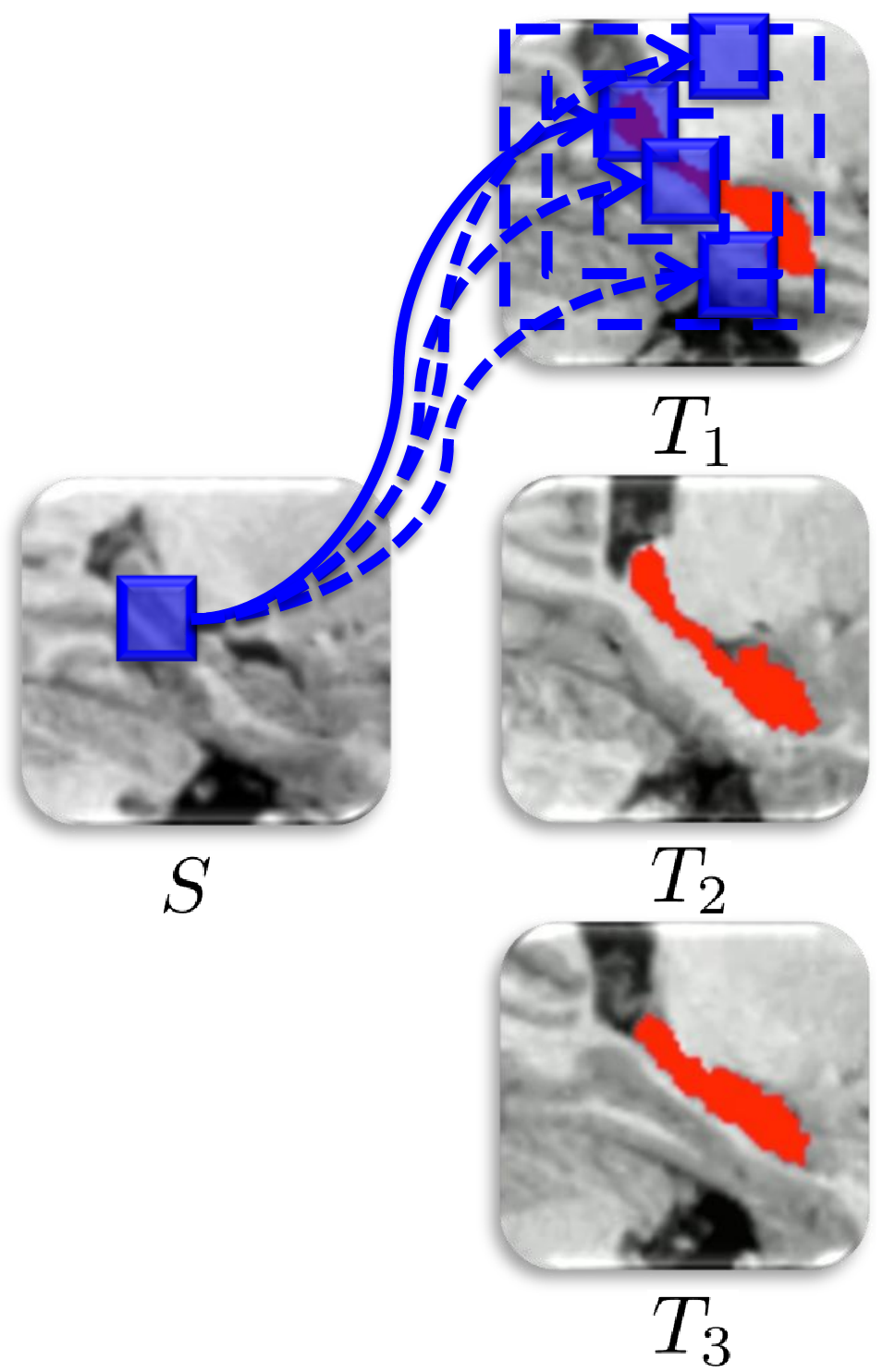}}}  
  \subfigure[multiple OPM]
  {\label{subfig:mp}\fbox{\includegraphics[width=\siz,height=140pt]{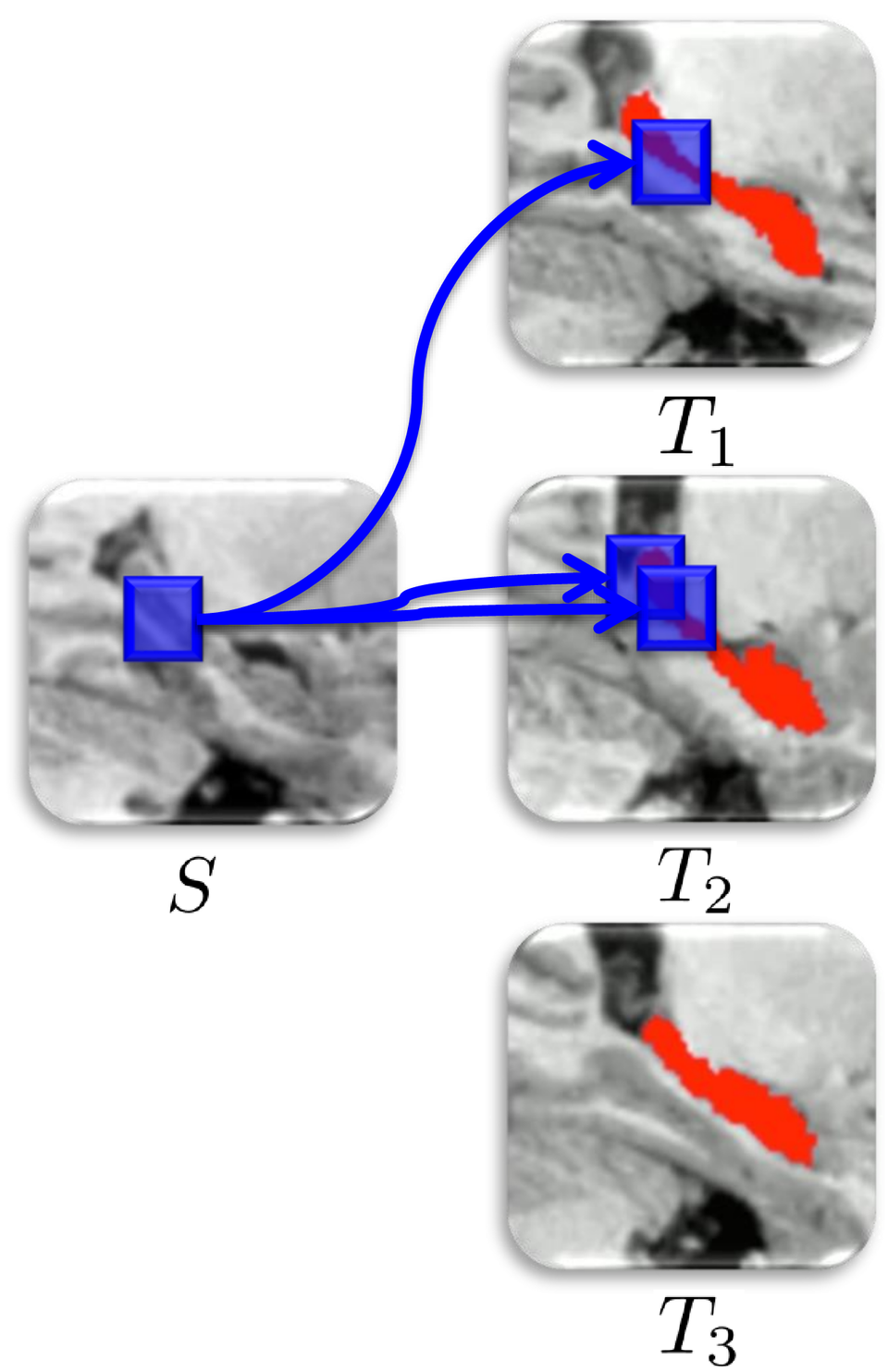}}}
}
\caption{Optimized PatchMatch (OPM) main steps.  
  In this figure, the representation of OPM steps focuses on the blue patch in $S$.
  Green, pink, purple and orange colors represent the adjacent patches of the blue patch. 
  During the constrained initialization (CI) \subref{subfig:init}, patches of the subject $S$ are matched (full lines) 
  to a random patch of the library within an initialization search window (three are displayed).
  The propagation step (PS), is represented for iteration \#1 and \#2  in \subref{subfig:prop:one} and \subref{subfig:prop:two}, respectively. 
  The shifted correspondences of recently processed adjacent patches are tested for improvement (dotted lines).
  Constrained random search (CRS) for iteration \#1 and \#2 are represented for the blue patch, in \subref{subfig:rs:one} and \subref{subfig:rs:two}, respectively. 
  Random tests are performed within a decaying search window around the current best match, within the current best template.
  In \subref{subfig:mp}, the result of multiple independent ANN searches by OPM is illustrated.
  See text for more details.
  }
\label{fig:opal}
\end{figure}

The OPAL algorithm is explained in detail in the next section and 
Figure~\ref{fig:opal} proposes a schematic overview. 
To clearly illustrate our Optimized PatchMatch (OPM) key steps, in Figure~\ref{fig:opal},
only three templates are considered as 
template library $T$, two iterations are performed 
and 3D MRI volumes are displayed in 2D.

As in the original paper, the metric used to compare the distance between a patch centered on $\mathbf{x_i}\in A$ and a
patch centered on $\mathbf{x_j}\in B$,
is a sum of squared differences (SSD),
\begin{equation}
dist(\mathbf{x_i},\mathbf{x_j})=\sum_{\sigma \in \Omega_s}^{}(A(\mathbf{x_i}+\sigma )-B(\mathbf{x_j}+\sigma ))^2 ,
\label{ssd}
\end{equation}
where $\Omega_s $ is the index coordinate set of the $s{\times}s$ 2D patch, centered on $(0,0)$,
considering $s$ as the patch size.

\subsubsection{\label{subsubsection:CI} Constrained Initialization} 
In the PM original paper \cite{barnes2009patchmatch}, the
initialization consists in assigning, for each patch located at
$(x,y)\in A$, a random
correspondence which can be located everywhere at $(x',y')\in  B$.
In the case of multi-templates method based on 3D MRI, the natural extension of this 
initialization step is to assign, for
each patch of the 3D image of the subject to segment $S$ located at $\mathbf{x_i} = (x,y,z) \in S$, 
a random patch correspondence
located at $\mathbf{x_j} =\{(x', y', z'), t\}$ where
$t\in\{1,\ldots,n\}$ is the index of the template $T_t$ within the template
library $T$. 
However, as we deal with linearly registered MRI volumes, we
propose to constrain the random initial position $(x',y',z')$ to be within
a fixed search window centered around the current voxel position
$(x,y,z)$. 
Then, for each voxel in $S$, an index template $t$ is assigned using
$i.i.d.$ random  variable within $\{1,\ldots,n\}$. 
Consequently, each patch in $S$ is associated to a unique random match among all templates of the library $T$.
Considering the important number of patches in $S$, all templates are very likely to be reached at least once.
Moreover, although the corresponding template is randomly selected during the initialization step,
all matches can move from a template to another during the following iterative process.
Figure~\ref{subfig:init} illustrates the initialization step. For each patch in $S$
(only three are displayed), the fixed search
window for the random initialization 
is depicted in dotted lines in the different training templates. 

This constraint has two advantages. First, it improves the matching convergence, 
making good use of the linear registration between training template  and the subject. 
Second, limiting the initialization to a fixed window prevents the algorithm from
finding similar patches in terms of intensity (low SSD) that are spatially far,
leading to potential segmentation errors. As a consequence, 
our constrain initialization reinforces spatial 
proximity between voxels in $S$ and their matches in $T$ 
and makes the algorithm converge faster.

As in the original PatchMatch algorithm, after this constrained initialization, 
propagation and random search steps are performed
iteratively in order to improve the patch correspondence.

\subsubsection{Propagation Step with Fast Distance Computation}
The propagation step of OPM is the 3D extension of the one proposed in
\cite{barnes2009patchmatch}.
For each patch located at $(x,y,z) \in S$, an ANN improvement is performed
by testing if the shifted ANN of its $6$ directly adjacent 
patches located at $(x\pm 1, y, z)$, 
$(x, y\pm 1, z)$ and $(x, y, z\pm 1)$ provides a better match. 
In order to converge faster and to propagate good correspondences, 
the original PM only tests recently processed neighbors during this step. 
Consequently, in 3D, only three adjacent neighbors are tested at each iteration,
according to the raw scan order.
Figures~\ref{subfig:prop:one} and~\ref{subfig:prop:two}
illustrate this step, where the blue
dotted lines 
correspond to the test of shifted adjacent neighbors in $T$, in order to
improve the current blue patch correspondence.
In this example, the best match for
the blue patch moves from template $T_1$ to $T_2$ with iteration $\#1$
and from $T_2$ to $T_1$ with iteration $\#2$.
The propagation step is a core stage of the OPAL algorithm since it allows a patch
correspondence to move over all the templates in $T$. 
Thus, the ANN of the current voxel can move
from one template to another one, since the ANN of the adjacent voxels are not
necessarily in the same template.

Moreover, the computational burden of these tests
can be extremely reduced in the
propagation step. Indeed, we propose an acceleration technique 
based on the observation that the ANN of the adjacent patches are known. 
As neighbor patches are overlapping, we use a shifted SSD instead of computing the whole distance between the current patch and the 
shifted ANN of its adjacent patch.
Hence, only the non overlapping coordinates are considered, {\emph i.e.}, 
the two squares at 3D patches extremities, 
since there is a one voxel shift in only one of the three dimensions.
The exact SSD between the current patch and the shifted correspondence is thus obtained in the fastest way.
The patch overlapping is illustrated in Figure~\ref{subfig:prop:one},
where the blue square overlaps the green and pink ones. 
The distances on the overlapping areas do not need to be re-computed.

\subsubsection{Constrained Random Search}
In the original PM algorithm \cite{barnes2009patchmatch}, 
the random search step is performed on all dimensions. 
In contrast to the original method, 
OPAL deals with a library of images. Therefore, we modify the 
random search step to take into account this aspect. 
In order to ensure spatial consistency, OPAL  performs
the random search only in the current template containing the 
current best patch correspondence (\emph{i.e.}, $t$ is
fixed, and we random on $(x_t', y_t', z_t') \in T_t$) 
within a search window decaying by a factor $2$.
The process stops when the window is reduced to a single voxel.
The decaying search window size is empirically defined as the size of the initialization window.
Figures~\ref{subfig:rs:one} presents
examples of such fixed template   
random search where the decaying search windows 
are represented in dotted blue lines.

\subsubsection{\label{subsubsection:MP} Multiple PM and Parallel Computation}
Contrary to \cite{barnes2009patchmatch} that only estimates the best match with PM,
OPAL computes $k$-ANN matches in $T$. These ANNs are then used
to perform the label fusion.    
In the literature, an extension of the original PM algorithm to $k$-ANN
case has been proposed in \cite{barnes2010generalized}. The suggested
strategy is to build a stack of the best visited matches. At each 
new tested match, the distance is compared to the one of the worst ANN among the stack.
If there is an improvement in terms of SSD, the worst ANN is replaced by the new match.
However, to parallelize such an approach, the current image $S$ must
be split into several parts. Since PM uses propagation 
of good matches between adjacent patches, 
any split would lead to boundary issues. 
Therefore, in OPAL, we decide to implement the $k$-ANN search through $k$ independent OPM, denoted as $k$-OPM.
This leads to a more efficient and simple multi-threading.  
Consequently, each thread can run an OPM without any
dependencies to the other ones.
Figure~\ref{subfig:mp} illustrates the result of the multiple OPM steps
with $k=3$.  One can note that $k$ independent OPM can lead to the same ANN for a given voxel.
The redundancy of the same ANN in the ANN map is not an issue, 
since each contribution is weighted 
during the patch-based label fusion step. During our validation, for the considered size of training libraries, 
we experimentally observed that such multiple selections of the same ANN is a rare phenomena.

\subsection{Patch-based Segmentation}

\begin{figure}[h!]
\centering
\centerline{
  \fbox{\includegraphics[width=300pt]{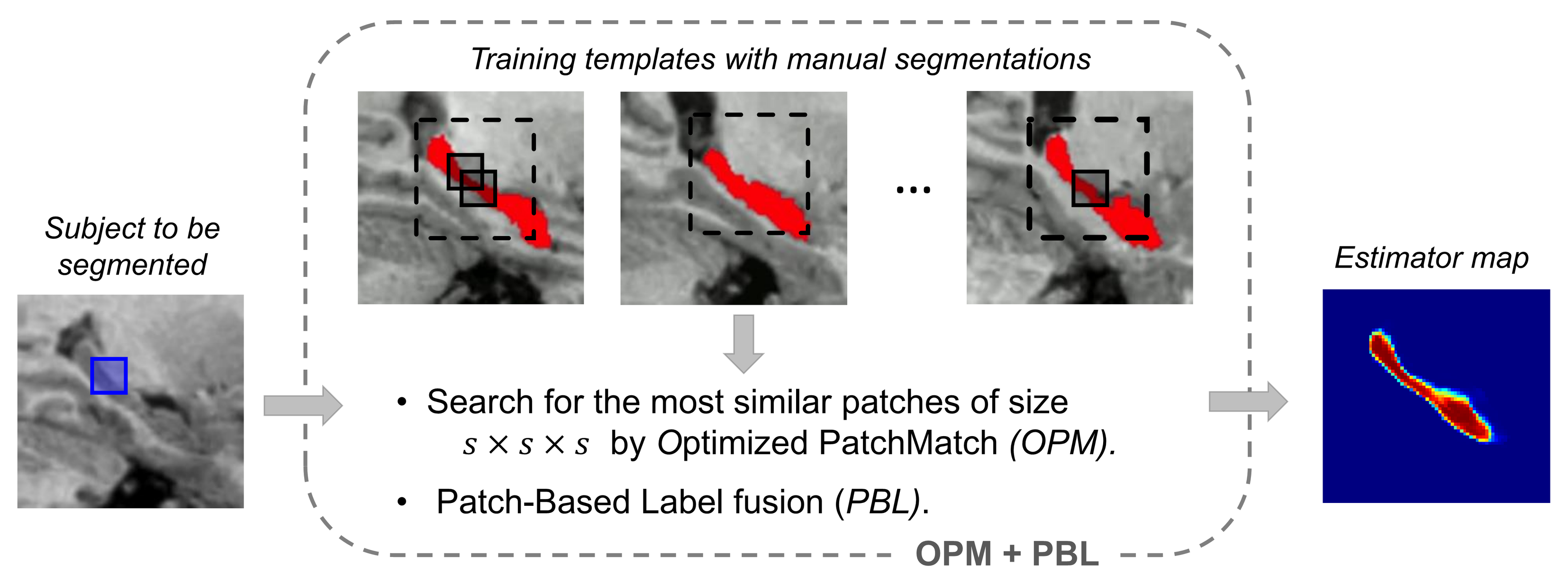}}
  }
\caption{Core of OPAL method: optimized PatchMatch and patch-based label fusion on image intensities.
For every voxel of the subject to segment, 
a search for similar patches of size $s{\times}s{\times}s$ is carried out by OPM.
A patch-based label fusion is then performed to generate a label estimator map.
See text for more details.}
\label{fig:method1}
\end{figure}

After convergence of the multiple OPM, the position and the distance of the $k$-ANN is known. 
Therefore, a patch-based label fusion step can be used to produce the final segmentation. 
In such a method, labels are fused according to their relevance to compute an estimator map of the subject to segment.  
In contrast to the original PBL method \cite{coupe2011patch}, 
where only the central voxel information was considered,
OPAL segmentation is performed in a patchwise manner, using the whole training patch as done in \cite{rousseau2011supervised, wu2013generative, manjon2014}. 
Moreover, as recently proposed in \cite{manjon2014}, OPAL uses a bilateral kernel for weight computation in order to reinforce spatial coherency.  
Figure~\ref{fig:method1} illustrates the patch-based label fusion process 
and the computation of the estimator map and is detailed below.

\subsubsection{\label{subsubsection:PBL} Patchwise Label Fusion}

At the end of the matching process, the $k$-ANN are estimated for all
the patches in $S$. Thus, the location and the SSD between the patches
of $S$ and their $k$-ANN in $T$ are known. 
To obtain the
final segmentation, we use the Patch-based label fusion (PBL) method presented in
\cite{coupe2011patch}. In contrast to \cite{coupe2011patch}, 
that considers all the patches within a fixed number of preselected templates,
OPAL only uses the $k$ most similar
patches (limiting segmentation error) over the entire library
(increasing segmentation accuracy). As previously mentioned, when the same ANN is
selected several times by independent PM, 
it will be taken into account several times during the label fusion.
Considering a 3D patch $\PP (\xii )$ at voxel position $\xii =(x,y,z)\in S$, 
and $\KK _i = \{\xjt \}$ the set of its $k$-ANN match positions, 
its label fusion $\LL (\xii )$ is defined by,
\begin{equation}
 \LL (\xii ) =\frac{\sum_{\xjt \in \KK _i}^{} \omega(\xii ,\xjt )l(\xjt )}
            {\sum_{\xjt \in \KK _i}^{} \omega(\xii ,\xjt )}   ,
\end{equation}
where $\omega(\xii ,\xjt )$ is the weight assigned to
$l(\xjt )$, the binary label given by the expert
at voxel $\xjt =\{\xjj ,t\}\in T$.

The weight $\omega(\xii ,\xjt )$  depends
on the similarity between the patches $\PP (\xii )\in S$, 
the patch contributing to the labeling of $\xii $,
and the ANN patch $\PP (\xjt )\in T$. This weight is defined as,
\begin{equation}
\omega(\xii ,\xjt ) = \exp{(1-\frac{\|\PP (\xii ) - \PP (\xjt )\|_2^2}{h(\xii )^2})} , 
\end{equation}
where $h(\xii )^2=\alpha ^2 \underset{\xjt \in \KK _i}{min}(\|\PP (\xii ) - \PP (\xjt )\|_2^2 + \epsilon ) $, 
with $\epsilon $ a small constant to ensure numerical stability, and $\alpha $ a normalization constant.
With the parameter $h(\xii)$, the distance of the current contribution is divided
by the minimal distance among all $k$-ANN contributions.

Most nonlocal label fusion methods perform voxelwise aggregation,
which can provide a lack of regularization on final segmentation.
Therefore, to further improve segmentation quality, the
label fusion is performed over the whole
patch as done in \cite{rousseau2011supervised,wu2013generative, manjon2014} and not
only using the central voxel. The patchwise labeling is then computed as follows,
\begin{equation}
 \LL (\PP (\xii )) =\frac{\sum_{\xjt \in \KK _i}^{} \omega(\xii ,\xjt )l(\PP (\xjt ))}
            {\sum_{\xjt \in \KK _i}^{} \omega(\xii ,\xjt )}   . \\
\label{lmap}
\end{equation}
This way, 3D patches $\PP (\xii )\in S$ are labeled at the same time. 
At the end, the label estimator for voxel $\xii $ is
obtained by averaging all neighbors contributions from overlapping blocks containing $\xii $ to obtain the estimator map  $\FF $.

\subsubsection{Bilateral Kernel}

In addition to the patchwise strategy,
a spatial filtering is performed during segmentation in order to 
reinforce spatial coherency of the selected $k$-ANN.
The spatial filtering exploits the observation that structures of interest are spatially 
close due to the linear registration. Therefore, good patch candidates should be similar in term of intensity and spatially not too far. Therefore, as done in NICE \cite{manjon2014},
each ANN contribution to patchwise labeling is also weighted by 
the spatial distance between patch centers $\xii \in S$ and $\xjt = \{\xjj ,t\}\in T$, 
\begin{equation}
\omega(\xii ,\xjt ) = \exp{(1-(\frac{\|\PP (\xii ) - \PP (\xjt )\|_2^2}{h(\xii )^2}
			    + \frac{\|\xii -\xjj \|_2}{\sigma ^2}))} , 
\label{weight}
\end{equation}
where $\sigma^2$ is a normalization constant.

\subsection{Late Aggregation of Multi-Scale and Multi-Feature Estimators}

\begin{figure}[h!]
\centering
  \centerline{
  \fbox{\includegraphics[width=340pt]{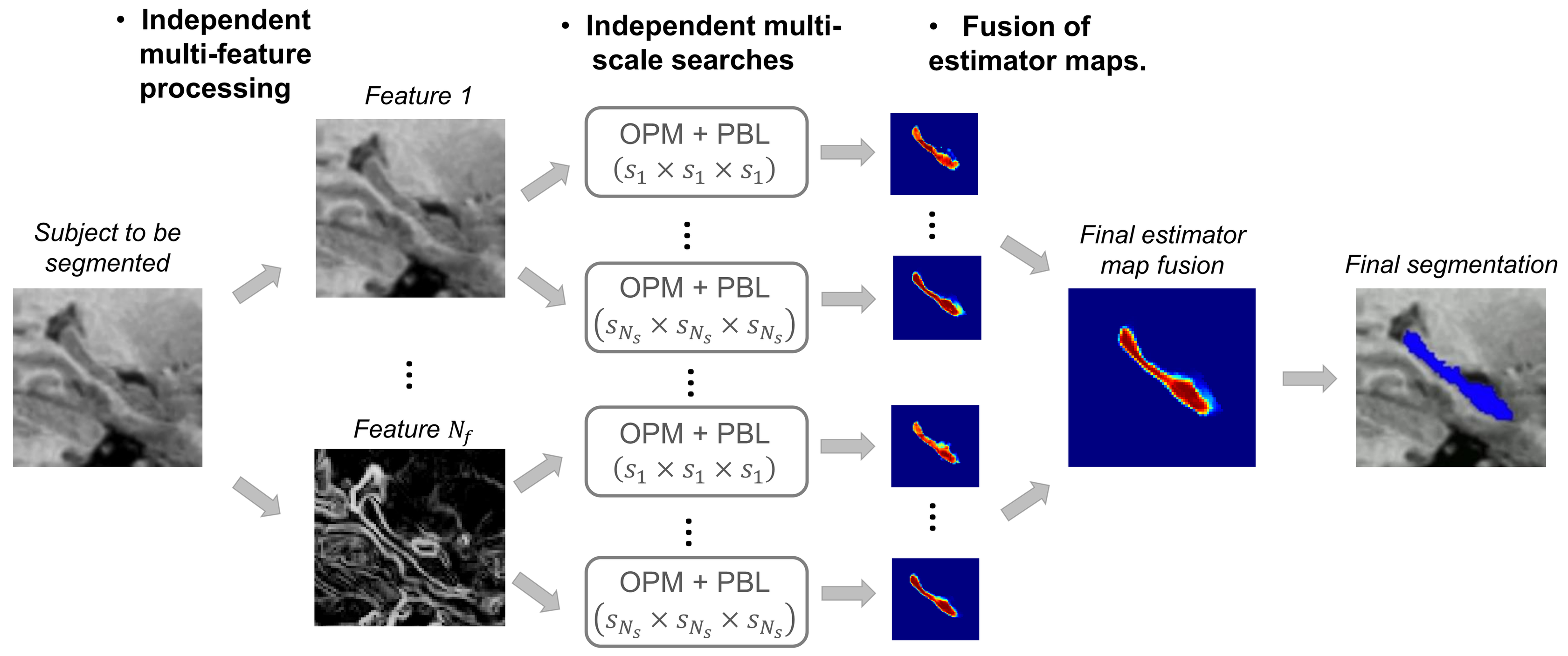}}
}
\caption{OPAL method. Fusion of multi-feature and multi-scale label estimator maps.
The algorithm is applied with $N_s$ different patch sizes, 
on $N_f$ different features, so $N=N_s{\times}N_f$ estimator maps are computed 
and merged to provide the final segmentation.
See text for more details.}
\label{fig:method2}
\end{figure}

Due to the high computational cost of previously published multi-templates methods, 
most were designed in a mono-scale and mono-feature context.
Recently, multi-scale \cite{eskildsen2012beast,wu2015,wachinger2014}, and multi-feature \cite{kim2013, bai2015}  
approaches have been investigated. These studies show the advantage of such frameworks.
However, since these methods require a non negligible computation time, they are based on either multi-scale \cite{eskildsen2012beast,wu2015,wachinger2014}
or multi-feature \cite{kim2013, bai2015} estimation but not both at the same time. 
Moreover, these methods perform early feature aggregation:
all the considered scales or features are fused into a single vector before performing patch comparison. 
However, early fusion is not necessarily the best strategy. Usually used for computation time consideration, 
early fusion has been shown to be less efficient than late estimator fusion/aggregation \cite{snoek2005early}. 
Moreover, the use of both multi-scale and multi-feature should improve segmentation accuracy. 
Leveraging the computational efficiency of OPAL, we propose to investigate a new framework to simultaneously perform
multi-scale and multi-feature analysis with late aggregation of estimators.
Figure~\ref{fig:method2} illustrates the whole OPAL method and the late fusion 
of multi-feature and multi-scale label estimator maps.

\subsubsection{\label{subsubsection:multiscale} Multi-scale Estimators}

In patch-based methods, the structure description highly depends on the size
of the patch. The patch size needs to be large enough to capture the
local geometry and to prevent discontinuities in the
segmentation. However, using very large neighborhoods may reduce the probability of finding similar patches in the library. 
Although the optimal patch size 
can be determined by experiments for a given dataset,
multi-scale approaches may significantly improve segmentation accuracy as shown in recent
multi-scale label fusion approaches \cite{wu2015,wachinger2014}.
In these papers, the ANN search consists in finding 
the candidate minimizing the distance for every scale at the same time. 
Therefore, such a strategy selects a consensual candidate providing the best similarity on average over all the considered scales. 
In contrast to these previous works,
we propose to perform fully independent multi-scale ANN searches where a candidate providing the best similarity is obtained for each scale.
With this method, $k$-OPM are independently computed for multiple patch sizes
$s_i, i\in \{1,\dots ,N_s\}$. 
Consequently, in our context, multi-scale refers to the simultaneous use of patches of different sizes,
and the images are considered with their initial resolution.
In Figure~\ref{fig:method2}, the ANN search by OPM and PBL are performed on each feature 
for $N_s$ patch sizes.

\subsubsection{\label{subsubsection:multifeature} Multi-feature Estimators}

Similarly, the search for similar patches by OPM can also be
carried out independently on different features (edges, textures, etc.). 
During our tests with different potential features, we found  that using the 
gradient norm (\emph{i.e.}, first intensity derivative) in addition to the original MRI intensities increases the segmentation accuracy. 
Therefore, we use both these features. Figure~\ref{fig:method2} shows how OPAL
is applied to the $N_f$ features extracted from the subject $S$ to segment. 
The resulting estimator maps are then merged \emph{a posteriori} as explained in the next section. 
As for the multi-scale aspect, our framework contrasts with recent multi-feature methods \cite{bai2015} 
where the ANN search consists in  finding the best candidate for every feature at the same time.
In our method, the independent searches improve  the ANN diversity of the selected matches.

\subsubsection{Late Aggregation of Estimators}

Label estimator maps are independently computed from PBL on 
multi-scale and multi-feature ANN searches.
The last step is the aggregation of these estimator maps to generate the final segmentation. 
Here, OPAL is applied on $N_f$ features, 
with $N_s$ different patch sizes,
so $N=N_s{\times}N_f$  estimator maps $\FF ^i$ with $i\in \{1,\dots ,N\}$
are computed to generate the final segmentation.
The final estimator map $\FF $ is then computed by 
averaging the estimator maps by a late fusion \cite{snoek2005early},
\begin{equation}
\FF = \frac{\sum_{i=1}^{N}\FF ^i}{N} .
\label{rfmaps}
\end{equation}

In the end, the final label decision is taken as follows:
\begin{equation}
\SSS (\xii ) = 
\begin{cases}
    1,& \text{if } \FF (\xii ) \geq 0.5 ,\\
    0,              & \text{otherwise}.
\end{cases}
\label{finaldec}
\end{equation}

\section{Materials}

\subsection{\label{subsection:dataset}Datasets}
During our experiments on hippocampus segmentation, two different datasets have been considered. 
We used images from elderly adults
obtained from the Alzheimer’s Disease Neuroimaging Initiative (ADNI)
dataset \cite{jack2008alzheimer} and images from young adults obtained from the International Consortium for Brain Mapping  (ICBM) dataset \cite{mazziotta1995}. 
Our goal was to demonstrate the robustness of our OPAL framework using data from different sources with different preprocessing pipelines.

\noindent\textbf{EADC-ADNI.}
This dataset was used to evaluate the performance of our approach. 
The European Alzheimer's Disease Consortium and Alzheimer's Disease Neuroimaging Initiative (ADNI) Harmonized Protocol (HarP) 
is a Delphi definition of manual hippocampus segmentation from MRI that can be used to validate automated segmentation algorithms \cite{boccardi2014delphi}. 
The EADC-ADNI dataset is based on ADNI MRI scans \cite{jack2008alzheimer} which were acquired on General Electric, Philips, and Siemens scanners using a 
3D MPRAGE T1-w sequence as recommended by the MRI Core of the ADNI consortium. The ADNI acquisition protocol is based 
on sagittal 3D MP-RAGE sequence (TR=2400ms, minimum full TE, TI=1000ms, FOV=240mm, voxel size of 1.25${\times}$1.25${\times}$1.2mm\textsuperscript{3}). 
Images were then reconstructed at a voxel size of approximately 1${\times}$1${\times}$1.2mm\textsuperscript{3}. 
As part of the EADC-ADNI, 100 MRI of the ADNI dataset have been manually labeled according to 
the harmonized protocol and are freely available (\url{www.hippocampal-protocol.net}).  
The definition of the harmonized protocol has been designed to reduce inconsistencies of 
manual segmentation protocols as detailed in  \cite{boccardi2014delphi}. 
The mean Dice value for repeated manual segmentations between experts has 
been estimated to $89\%$ ($[88\%;92\%]$) according to \cite{tangaro2014}. 
All the images were preprocessed using the volBrain pipeline (\url{http://volbrain.upv.es}).
The first preprocessing step is based on the adaptive nonlocal mean filter \cite{manjon2010adaptive}.
Denoised MRI are then coarsely corrected for inhomogeneity with N4 \cite{tustison2010n4itk}.
Afterwards, an affine registration to MNI space is achieved using ANTS \cite{avants2011reproducible}.
In the MNI space, a fine inhomogeneity correction is performed using SPM8 routines \cite{weiskopf2011unified}.
Finally, an intensity normalization procedure is applied to the images \cite{manjon2008robust}.
The whole preprocessing pipeline is performed in less than $5$min per subject.

\noindent\textbf{ICBM.}
We used a part of the International Consortium for
Brain Mapping (ICBM) dataset \cite{mazziotta1995}
which consists of 80  MR images of young and healthy
individuals with manual 
segmentations following the Pruessner's protocol
\cite{pruessner2000volumetry}.
The MRI scans were acquired with a 1.5T Philips
GyroScan imaging system (1mm thick slices, TR=17ms, TE=10ms,
flip angle=$30\,^{\circ}$, FOV=256mm).
The
estimated intra-class reliability coefficient was of $90\%$ for
inter- (4 raters) and $92\%$ for intra-rater (5 repeats) reliability.  
All the
images were preprocessed through the following pi\-pe\-li\-ne: estimation of
the standard deviation of noise 
\cite{coupe2010robust}; denoising using the optimized nonlocal means
filter \cite{coupe2008optimized};
correction of inhomogeneities using N3 \cite{sled1998nonparametric};
registration to 
stereotaxic space based on a linear transform to the ICBM152 template
($1{\times}1{\times}1\text{mm}^3$ voxel size)
\cite{collins1994automatic}; linear 
intensity normalization of 
each subject on template intensity; 
image cropping around the structures of interest; and
cross-normalization of the MRI intensity between the subjects 
with \cite{manjon2008robust}.
As for EADC-ADNI preprocessing, the whole pipeline requires less than $5$min per subject.

\subsection{Quality Metric and Compared Methods}
The proposed method was validated through a leave-one-out cross 
validation procedure for both datasets.
The segmentation accuracy was estimated with the standard
Dice coefficient (also called kappa index) introduced in \cite{zijdenbos1994}
 which compares the expert-based 
segmentation with the automatic segmentation.
 For two binary segmentations $\SSS _1 $ and $\SSS _2$, the Dice coefficient $D$ is computed as,
\begin{equation}
D(\SSS _1,\SSS _2)=\frac{2\mid \SSS _1\cap \SSS _2\mid}{\mid \SSS _1\mid +\mid \SSS _2\mid}  .
\end{equation}

 For each subject, the Dice coefficient of left and right hippocampus are averaged 
and the values in Tables~\ref{table:contribICBM},~\ref{table:contribADNI} and~\ref{tab:compICBM} 
correspond to 
the median Dice over all the dataset. The associated computation times include 
ANN map computation for every feature with every patch size, 
PBL on every estimator map and final
segmentation of both left and right hippocampus. 
During our validation process, we investigated the impact of 
parameters such as the initialization search window size, the patch size,
the number of neighbors (\emph{i.e.}, number of OPM), and the 
impact of multi-scale and multi-feature approaches on segmentation accuracy and computation time.

The results obtained by OPAL were compared to the published results on the ICBM dataset
of the original
Patch-Based Label fusion method (PBL)~\cite{coupe2011patch}, 
a Sparse Representation Classification method (SRC)~\cite{tong2013segmentation},
and a dictionary learning method, denoted as
Discriminative Dictionary Learning for Segmentation (DDLS)~\cite{tong2013segmentation}.
Mean Dice coefficients of left and right hippocampus results of 
EADC-ADNI dataset were compared to the results obtained with
a Random Forest approach~\cite{tangaro2014}, 
and two multi-templates based approaches,
BioClinica Multi-Atlas Segmentation algorithm (BMAS)~\cite{roche2014},
and Learning Embeddings for Atlas Propagation (LEAP)~\cite{gray2014}.

\subsection{Implementation Details}

OPAL was implemented in MATLAB using multi-threaded C-MEX code. Our
experiments were carried out using 
a server of 16 cores at 2.6 GHz
with 100 GB of RAM. 
Default parameters are set to process both ICBM and ADNI datasets.
These parameters offer a good trade-off between segmentation accuracy and computation time.
In the following results, 
OPAL is processed with 3 inner iterations of OPM and
 the number of threads on each feature is equal to $k$.
In~\eqref{weight}, parameters $\alpha $ and $\sigma $ are empirically set to 2.
In the multi-feature setting, estimator maps are computed from
 image intensities and gradient norm intensities.
In the multi-scale setting, OPAL is processed with
 $3{\times}3{\times}3$ and $5{\times}5{\times}5$ voxels
patch sizes on each feature.
Finally, the number of selected matches per voxel for 
each estimator is by default set to $k=10$ ANNs, and
the size of the initialization 
search window is set to $13{\times}13{\times}13$ voxels.
%

%

\section{Results}

\subsection{Influence of Parameters}
\begin{figure}[h!]
\centering
\newcommand{\siz}{0.49\textwidth}
\includegraphics[width=\siz, height=100pt]{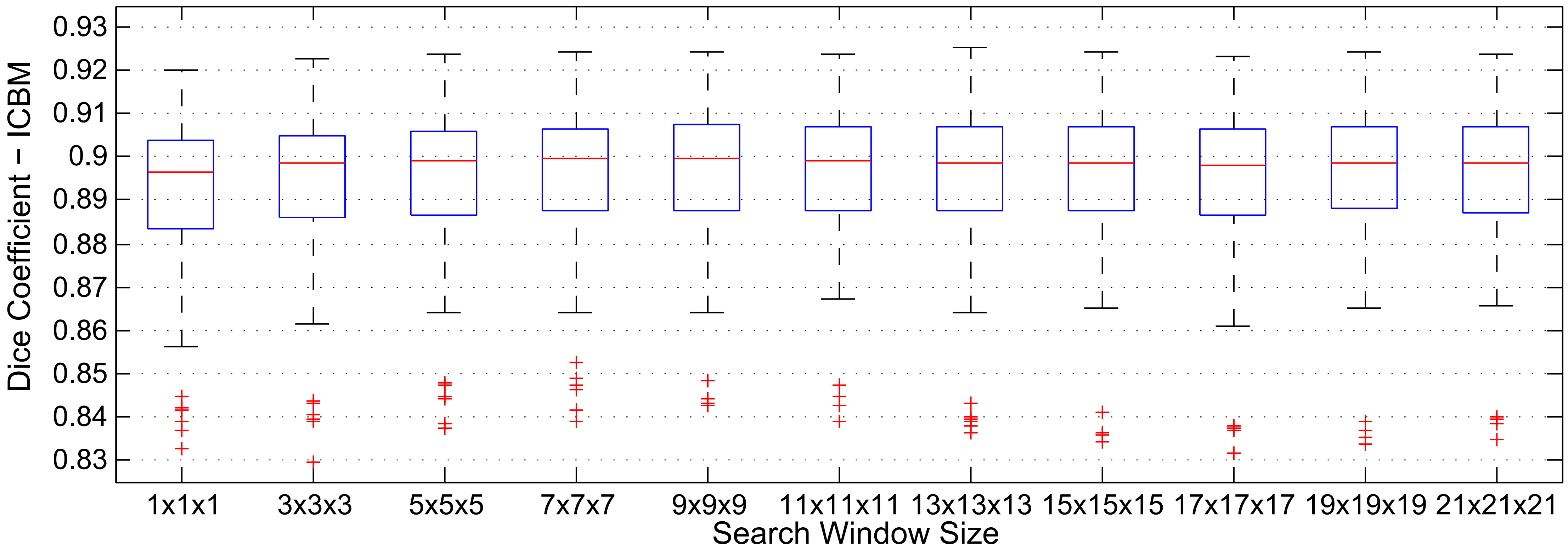}
\includegraphics[width=\siz, height=100pt]{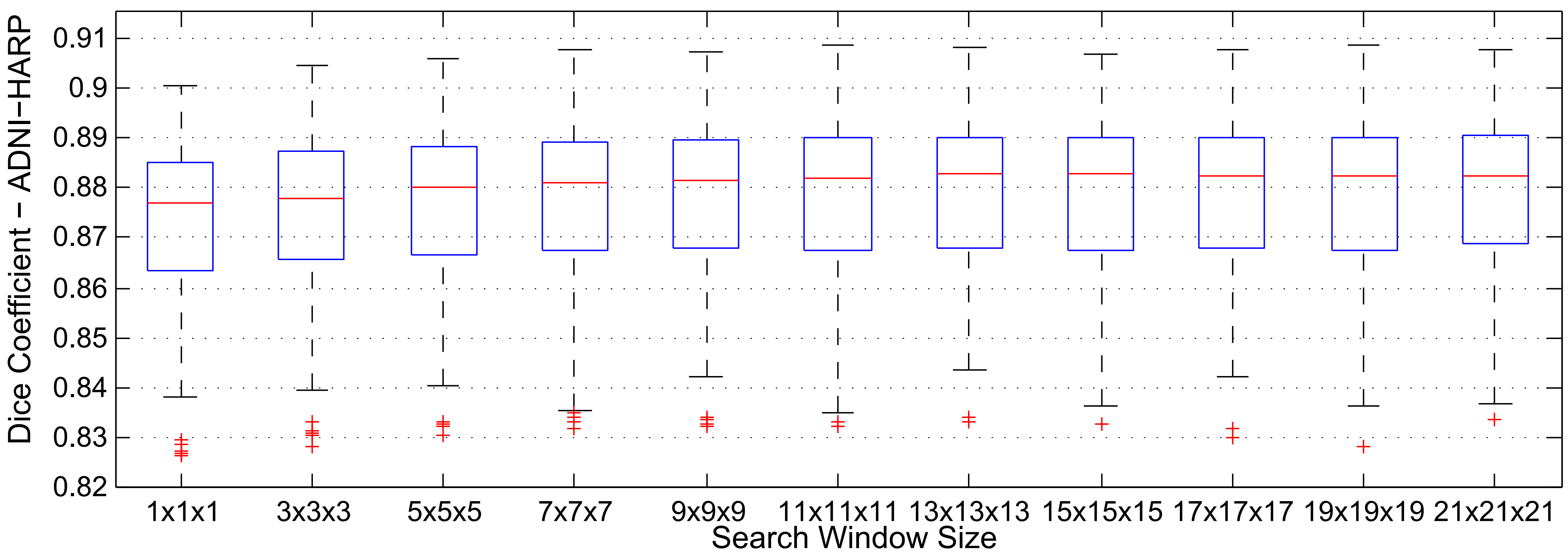}
\caption{Influence of the initialization search window 
on Dice coefficient for the ICBM (left) and the EADC-ADNI (right) datasets.} 
\label{fig:CIsize}
\end{figure}

First, as mentioned in~\ref{subsubsection:CI}, the initialization search window
reinforces spatial coherency
between voxels in $S$ and their matches in $T$. 
By setting the optimal search window area, the algorithm converges
faster since more relevant matches are found, thus
leading to a higher segmentation accuracy. 
This optimal window size is empirically estimated according to the dataset.
Figure~\ref{fig:CIsize} shows the Dice coefficient
for several initialization window sizes on both studied datasets.
For ICBM, a plateau is reached for a search window of $7{\times}7{\times}7$ voxels,
while an area of $13{\times}13{\times}13$ voxels leads to better 
segmentation results for the EADC-ADNI dataset. This second dataset requires a larger search window size since it 
contains higher anatomical variability due to the presence of pathologies.
Therefore, in the following, the initialization window is by default set to $13{\times}13{\times}13$ voxels.

\begin{figure}[h!]
\centering
\newcommand{\siz}{0.48\textwidth}
\newcommand{\sizz}{120pt}
\includegraphics[width=\siz, height=122pt]{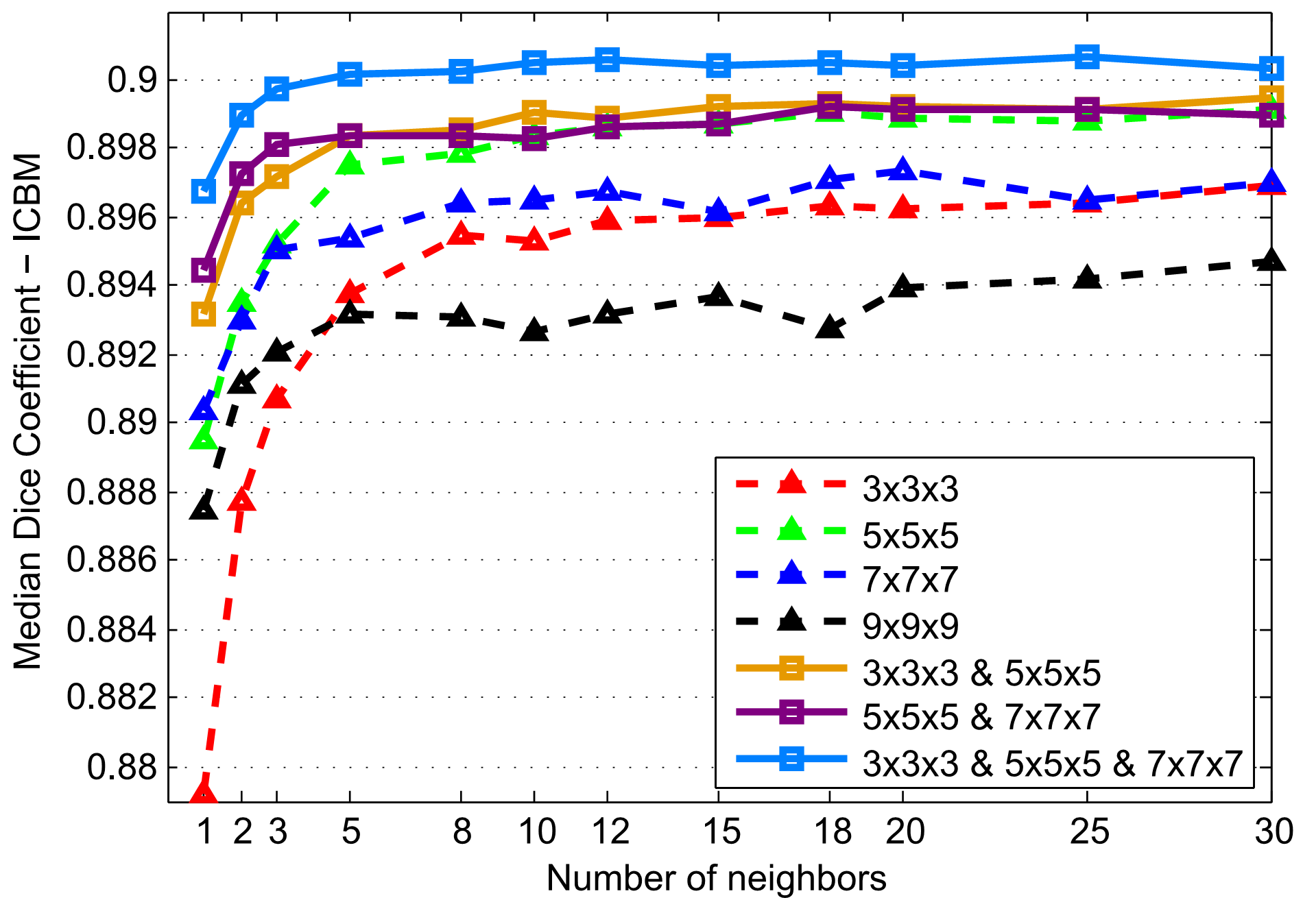}
\includegraphics[width=\siz, height=\sizz]{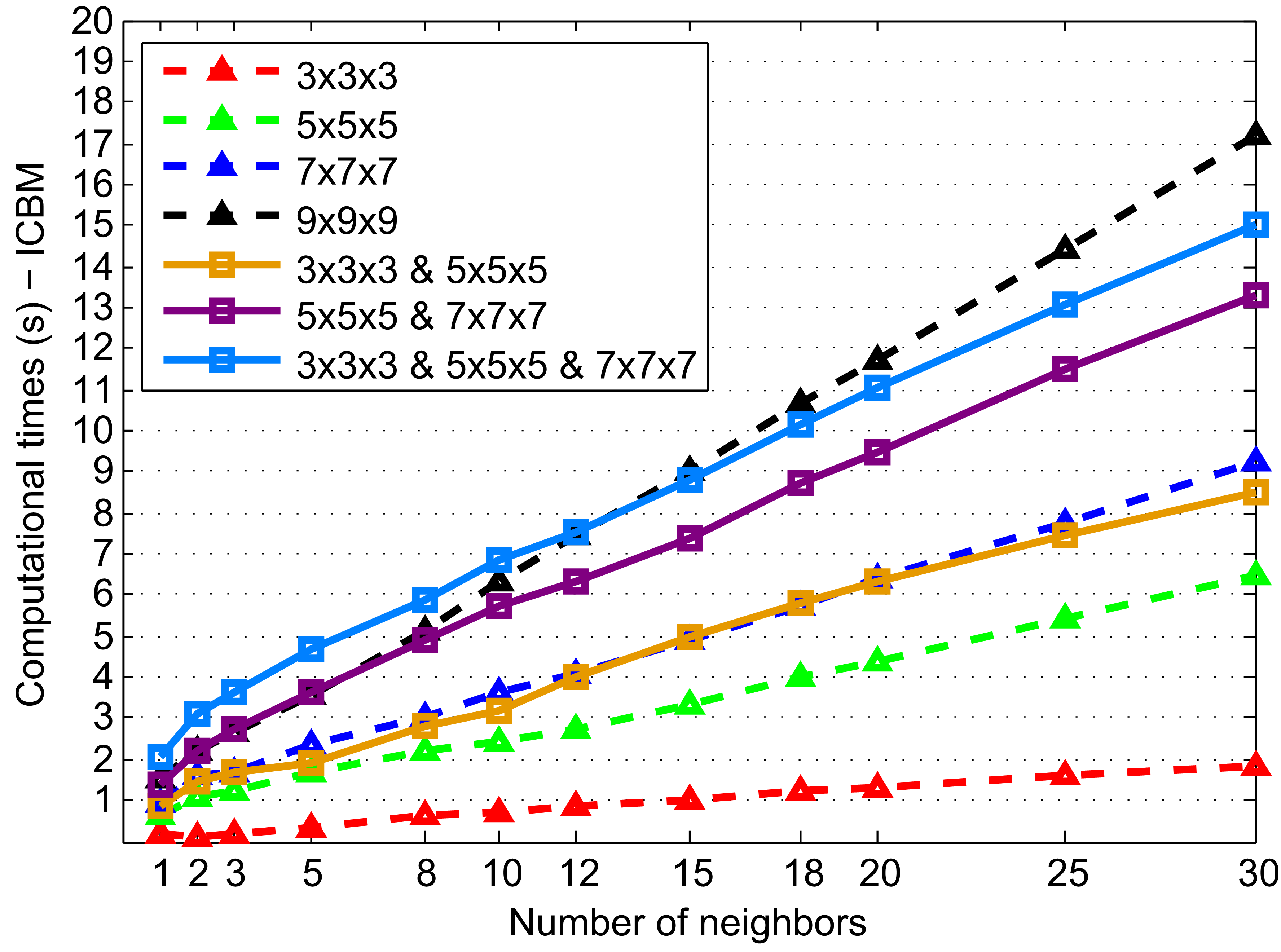}
\caption{
Median Dice coefficient according to the mono-scale and multi-scale patch sizes
    and the number of neighbors (left), and the corresponding
    computation time (right) for the ICBM dataset.
    These results are obtained with default multi-feature settings,
    \emph{i.e.}, MRI gradient norm in addition to the original MRI intensities.} 
\label{fig:knn_vs_ps_ICBM_kappa}
\end{figure}

\begin{figure}[h!]
\centering
\newcommand{\siz}{0.48\textwidth}
\newcommand{\sizz}{120pt}
\includegraphics[width=\siz, height=\sizz]{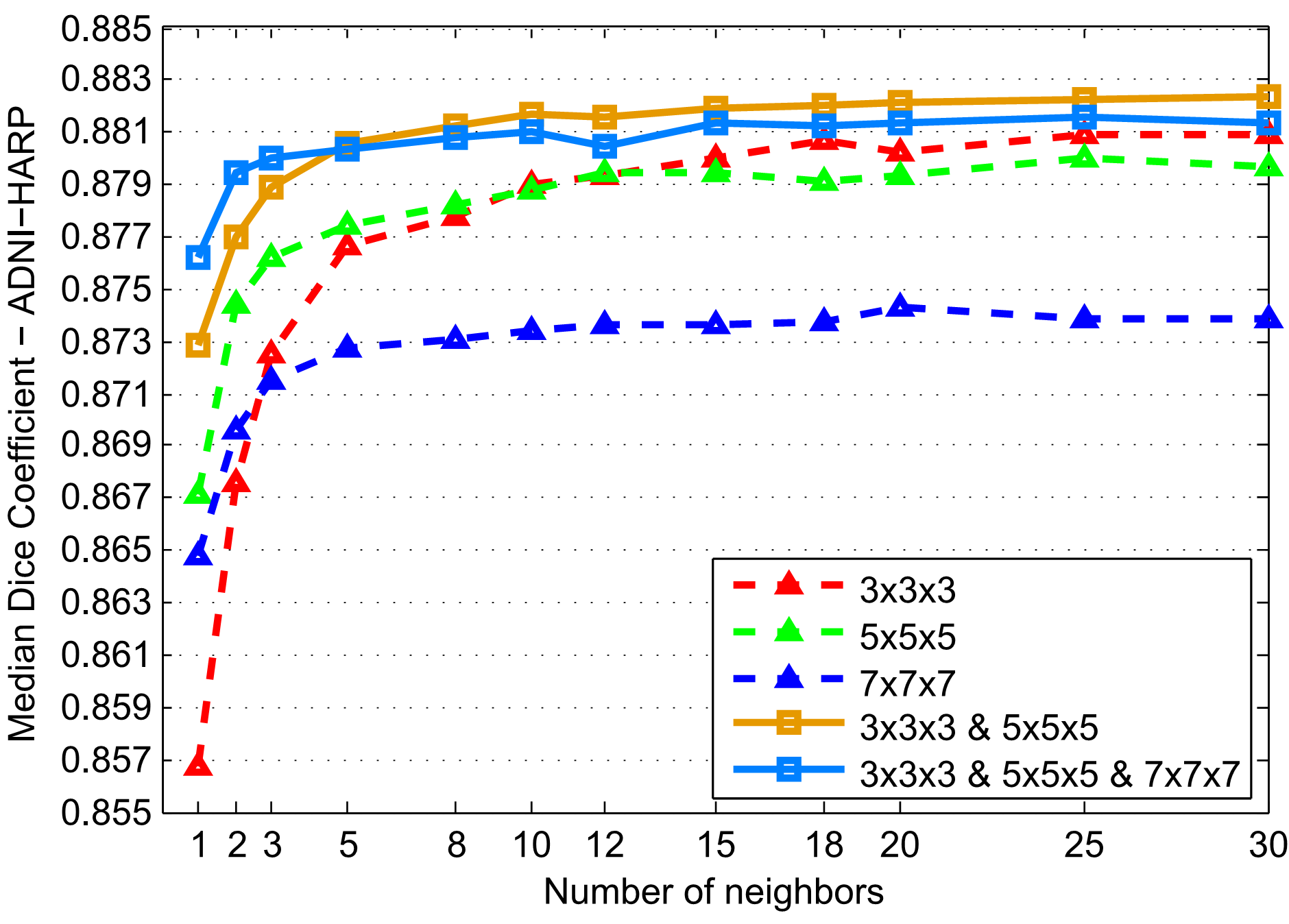}
\includegraphics[width=\siz, height=122pt]{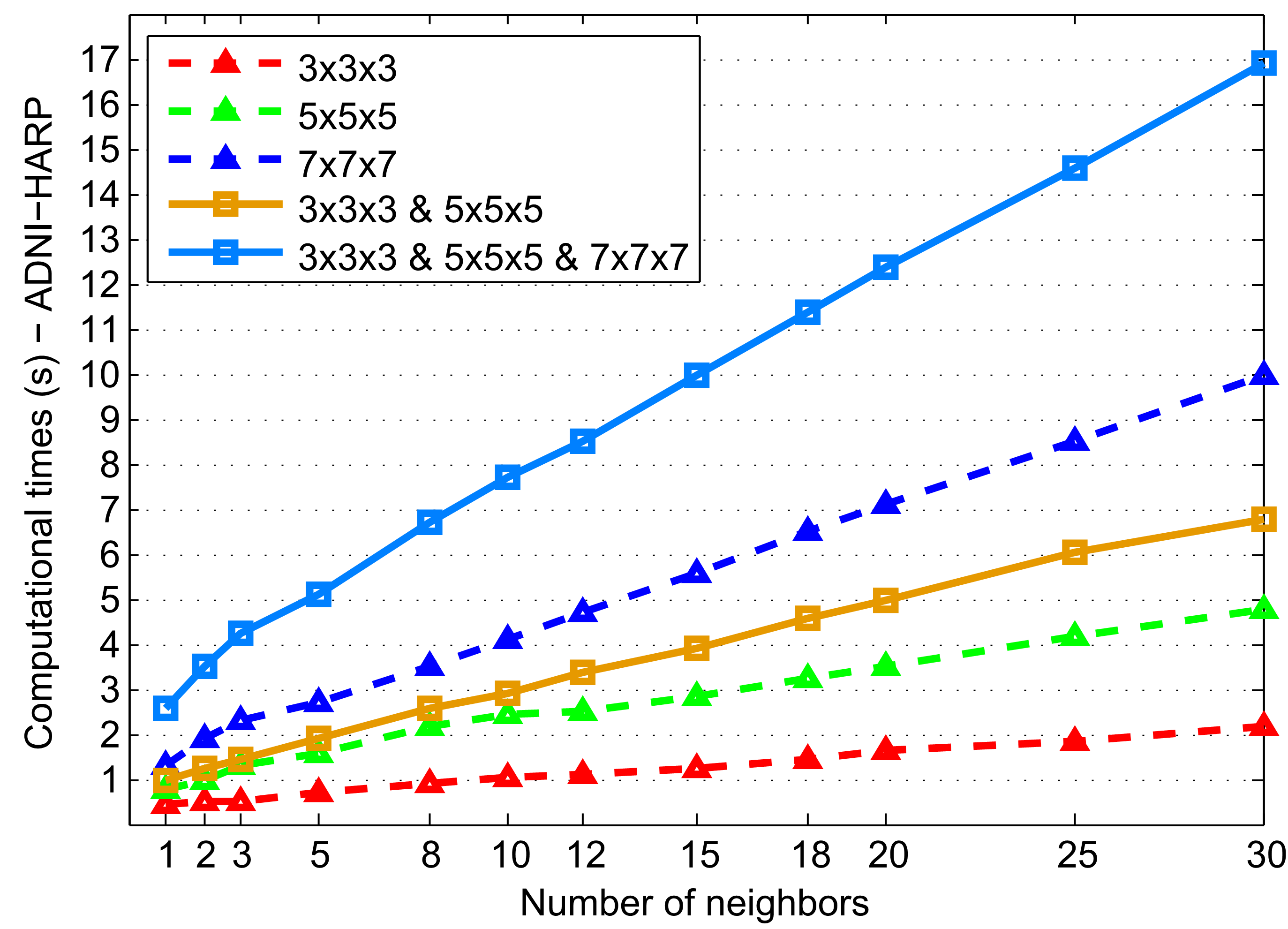}
\caption{
Median Dice coefficient according to the mono-scale and multi-scale patch sizes
    and the number of neighbors (left), and the corresponding
    computation time (right) for the EADC-ADNI dataset.
    These results are obtained with default multi-feature settings,
    \emph{i.e.}, MRI gradient norm in addition to the original MRI intensities.} 
\label{fig:knn_vs_ps_ADNI_kappa}
\end{figure}

Figures~\ref{fig:knn_vs_ps_ICBM_kappa} and~\ref{fig:knn_vs_ps_ADNI_kappa} 
show the influence of the number of
ANN (\emph{i.e.}, $k$) and of the patch size on the segmentation quality and on the
computation time.
Without the multi-scale approach, we found out that patches of size $5{\times}5{\times}5$ voxels
provide the best results on both datasets.
This patch size indeed gives acceptable description for structures of different scales, as already observed in 
\cite{coupe2011patch, tong2013segmentation}.
With our multi-scale approach, we can automatically take advantage 
of different patch sizes that provide better results.
By merging estimator maps generated from $3{\times}3{\times}3$ and $5{\times}5{\times}5$ voxels
patch sizes, we reach a Dice coefficient of $89.9\%$ for the ICBM dataset, with default settings.
(\emph{i.e.}, $k$=10 ANNs, multi-scale, multi-feature and initialization window set to $13{\times}13{\times}13$ voxels). 
By adding estimator maps from $7{\times}7{\times}7$ voxels patch sizes and increasing the number of $k$-OPM,
we even reach a $90.1\%$ Dice coefficient.
For the EADC-ADNI dataset, we reach a 
$90.1\%$ Dice coefficient ($90.05\%$ with default parameters). 
For both datasets, the segmentation step is performed in less than $2$s of processing per subject.
These results highlight the importance of taking into account 
the diversity of information obtained from various patch sizes.
We noted that the
median Dice coefficient reaches a plateau around
$10$-ANN. It is interesting to note that this number is coherent with the suggested
number of templates in multi-template matching methods
\cite{collins2010towards}.  
As expected, bigger patches and larger number of ANN require higher
computation time. 
Consequently, our experiments suggest that using $k=10$ ANNs on each feature 
offers a good trade-off between
segmentation accuracy and computation time.

Different settings were compared using paired t-test on Dice coefficients.
The results in Tables~\ref{table:contribICBM} and~\ref{table:contribADNI} 
present the impact of each contribution on 
Dice coefficient and computation time during the segmentation process.
For both datasets, the use of multi-feature and multi-scale significantly improved the segmentation accuracy compared to mono-scale and mono-feature method, as assessed by $p$-values.
Moreover, in all studied cases, multi-scale and multi-feature approaches improved results of mono-scale and multi-feature method.
This demonstrates the complementary nature of the multi-feature and multi-scale strategy.

Estimator maps for several features and several patch sizes
are shown in Figure~\ref{fig:estimator_maps}, for a subject of the EADC-ADNI dataset. 
First, bigger patch sizes produce smoother estimator maps. Smaller patches are able to better capture finer details at the expense of noisier estimator maps. 
Second, the estimators based on gradient norm better define edge structure but are less robust to noise. 
Finally, the aggregation is able to produce a good trade-off between considered scales and features. 

Figure~\ref{fig:S3D} presents segmentation results of best, median and worst subjects obtained on the EADC-ADNI dataset.
First, we can see that automatic method produces a smoother segmentation than expert.
The patchwise label fusion obtains consistent segmentation along the edge, 
but tends to fill holes present in manual segmentation. 
Some of these holes appear to be hippocampal CSF while others seem to be expert inaccuracies.

\begin{table}[h!]
\caption{Influence of multi-scale and multi-feature
  in terms of segmentation accuracy and
  computation time on the ICBM dataset. 
  Mono-scale and mono-feature results are obtained with PBL from
  $5{\times}5{\times}5$ voxels patch size ANN search on MRI intensities.
  Multi-feature considers the MRI gradient norm 
  in addition to the original MRI intensities. 
  Multi-scale adds estimator maps computed from 
  $3{\times}3{\times}3$ voxels patch sizes
  on each feature. 
  The given computation times correspond to the mean segmentation processing time of one subject.
  }
  {\footnotesize
\label{table:contribICBM}
\setlength\belowcaptionskip{1pt}
\centering
\newcommand{\sz}{\hspace{8pt}}
\newcommand{\szz}{\hspace{2pt}}
\begin{tabular}{p{3.8cm}@{\szz}c@{\sz}c@{\sz}c@{\sz}c@{\sz}c@{\szz}}
\hline
OPAL on ICBM&Median Dice&Mean Dice&$p$-value&Comp. Time\\
\hline
Mono-scale, Mono-feature& $89.4\%$ &$89.1\pm1.85\%$&$<10^{-14}$&$0.27s$\\
+ Multi-feature&$89.8\%$&$89.6\pm1.68\%$&$0.0131$&$0.53s$\\ 
+ Multi-scale& $89.9\%$&$89.7\pm1.70\%$&${\times}$&$0.92s$\\
\hline
\end{tabular}
}
\end{table}

\begin{table}[h!]
\caption{Influence of multi-scale and multi-feature
  in terms of segmentation accuracy and
  computation time on EADC-ADNI dataset.
  Mono-scale and mono-feature results are obtained with PBL from
  $5{\times}5{\times}5$ voxels patch size ANN search on MRI intensities.
  Multi-feature considers the MRI gradient norm 
  in addition to the original MRI intensities. 
  Multi-scale adds estimator maps computed from 
  $3{\times}3{\times}3$ voxels patch size
  on each feature.
  The given computation times correspond to the mean segmentation processing time of one subject.
}
{\footnotesize
\label{table:contribADNI}
\setlength\belowcaptionskip{1pt}
\centering
\newcommand{\sz}{\hspace{8pt}}
\newcommand{\szz}{\hspace{2pt}}
\begin{tabular}{p{3.8cm}@{\szz}c@{\sz}c@{\sz}c@{\sz}c@{\sz}c@{\szz}}
\hline
OPAL on EADC-ADNI&Median Dice&Mean Dice&$p$-value&Comp. Time\\
\hline
Mono-scale, Mono-feature& $89.4\%$ &$89.2\pm1.55\%$&$<10^{-25}$&$0.49s$\\
+ Multi-feature&$89.7\%$&$89.6\pm1.45\%$&$<10^{-8}$&$0.95s$\\ 
+ Multi-scale& $90.1\%$&$89.8\pm1.46\%$&${\times}$&$1.51s$\\
\hline
\end{tabular}
}
\end{table}

\newcolumntype{M}[1]{>{\centering\arraybackslash}m{3.05cm}}
\newcolumntype{P}[1]{>{\centering\arraybackslash}m{2.2cm}}
\newcommand{\siztabb}{80pt}
\newcommand{\sizhb}{65pt}
\begin{figure}[h!]
\begin{table}[H]
\centering
\begin{tabular}{|P||M|M||}
\hline
\bf{Feature}&\multicolumn{2}{c|}{\bf{Patch size}}  \\ 
\hline
&$3{\times}3{\times}3$ & $5{\times}5{\times}5$  \\ 
MRI intensity&
\vspace{0.15\baselineskip} \includegraphics[width=\siztabb, height=\sizhb]{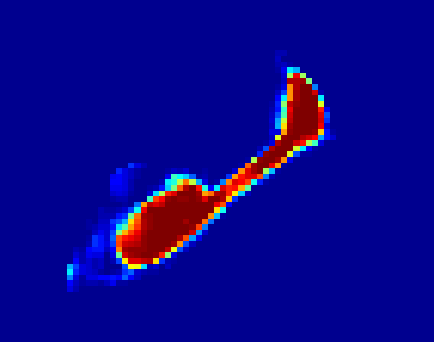}&
\vspace{0.15\baselineskip} \includegraphics[width=\siztabb, height=\sizhb]{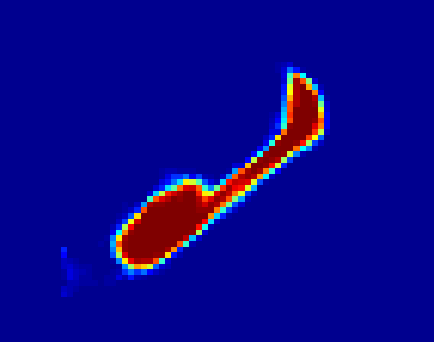}\\
MRI gradient norm&
\includegraphics[width=\siztabb, height=\sizhb]{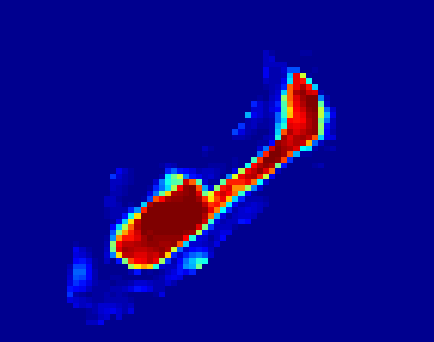}&
\includegraphics[width=\siztabb, height=\sizhb]{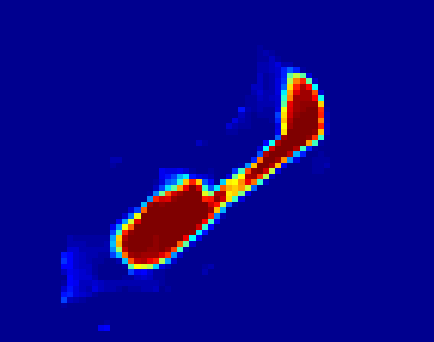} \\
Final map \& segmentation&
\vspace{0.35\baselineskip} \includegraphics[width=\siztabb, height=\sizhb]{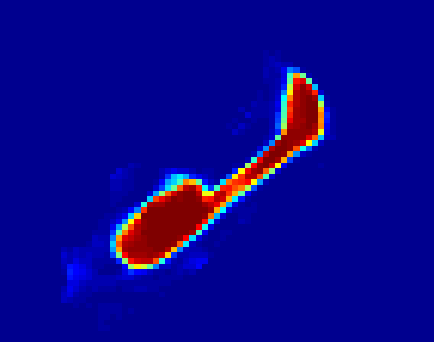}&
\vspace{0.35\baselineskip} \includegraphics[width=\siztabb, height=\sizhb]{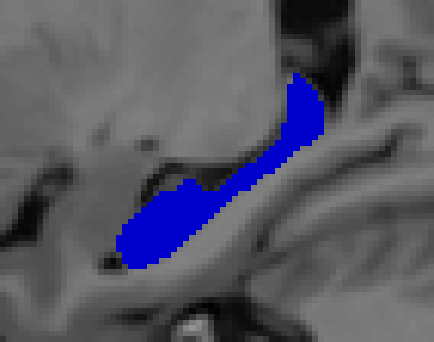}\\
\hline
\end{tabular} 
\end{table}
 \caption{2D visualizations of 
estimator maps for several features and several patch sizes
for the EADC-ADNI dataset.
With patches of size $5{\times}5{\times}5$ voxels, estimator map decision is more stable for every voxel
 (higher intensity within the hippocampus volume). With patches of size $3{\times}3{\times}3$ voxels, 
 some areas are more accurately
 segmented, see for instance the peak on top on the hippocampus image.
}
\label{fig:estimator_maps}
\end{figure}

\newcolumntype{M}[1]{>{\centering\arraybackslash}m{2.65cm}}
\newcommand{\siztab}{77pt}
\newcommand{\sizh}{63pt}
\begin{figure}[h!]
\begin{table}[H]
\centering
\begin{tabular}{|p{1.6cm}|M|M|M||}

\cline{2-4} \multicolumn{1}{c|}{ }&Best subject&Median subject&Worst subject \\ 
\cline{1-4} \multicolumn{1}{|c|}{ }&Dice=$92.4\%$&Dice=$90.1\%$&Dice=$85.8\%$\\
Expert 2D&\vspace{0.15\baselineskip} 
\includegraphics[width=\siztab, height=\sizh]{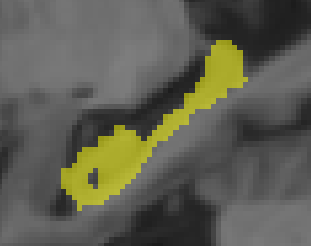}& 
\includegraphics[width=\siztab, height=\sizh]{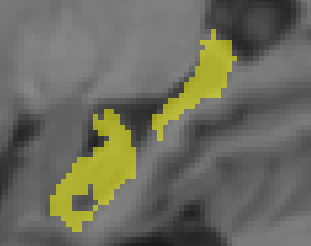}&
\includegraphics[width=\siztab, height=\sizh]{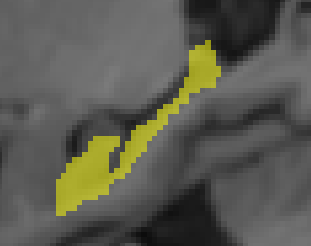}\\
Expert 3D&
\includegraphics[width=\siztab, height=\sizh]{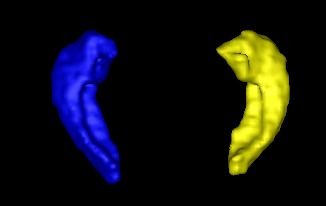}&
\includegraphics[width=\siztab, height=\sizh]{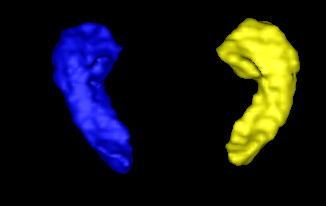}&
\includegraphics[width=\siztab, height=\sizh]{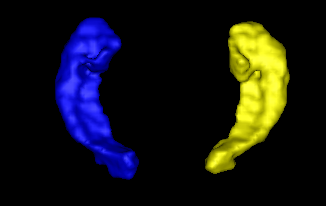}\\
OPAL 2D&
\includegraphics[width=\siztab, height=\sizh]{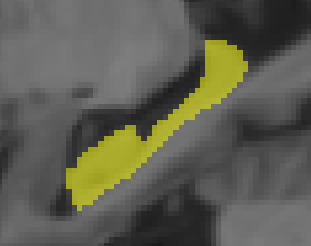}&
\includegraphics[width=\siztab, height=\sizh]{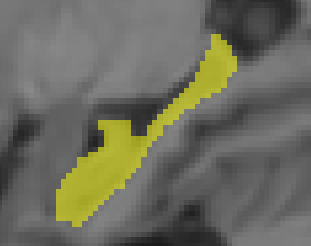}&
\includegraphics[width=\siztab, height=\sizh]{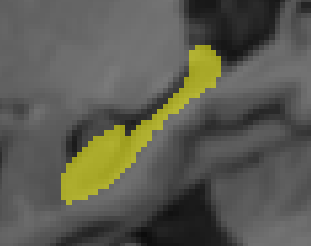} \\
OPAL 3D&
\includegraphics[width=\siztab, height=\sizh]{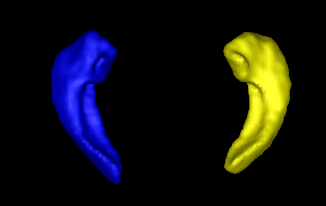}&
\includegraphics[width=\siztab, height=\sizh]{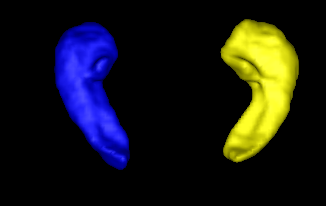}&
\includegraphics[width=\siztab, height=\sizh]{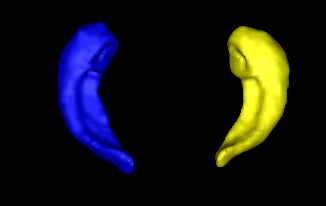} \\
Errors 2D&
\includegraphics[width=\siztab, height=\sizh]{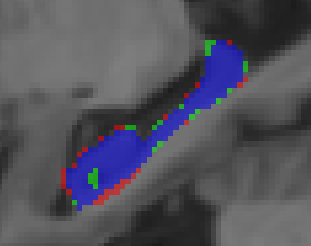}&
\includegraphics[width=\siztab, height=\sizh]{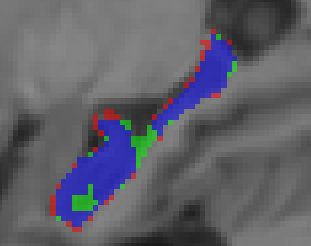}&
\includegraphics[width=\siztab, height=\sizh]{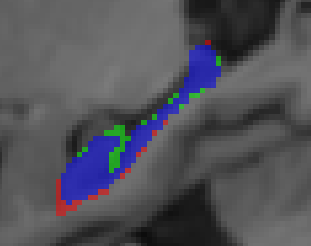}\\
Errors 3D&
\includegraphics[width=\siztab, height=\sizh]{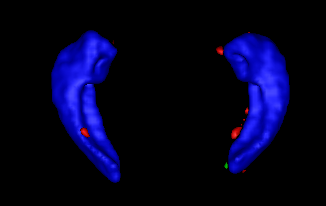}&
\includegraphics[width=\siztab, height=\sizh]{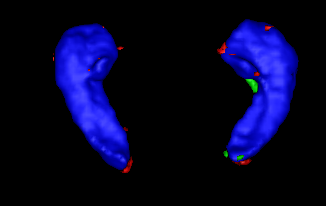}&
\includegraphics[width=\siztab, height=\sizh]{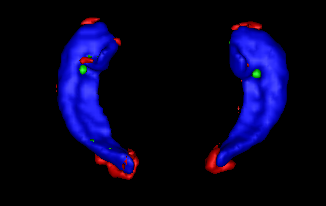}\tabularnewline
\hline
\end{tabular} 
\end{table}
 \caption{2D and 3D visualizations of best, 
 median and worst segmented EADC-ADNI subjects computed with default settings.
 In the fifth and sixth rows, blue voxels are overlapping with the expert segmentation, 
green voxels are the false positives (segmented by OPAL but not by the expert)
and red voxels are the false negatives (segmented by the expert but not by OPAL).}
\label{fig:S3D}
\end{figure}

\subsection{Comparison with State-of-the-Art Methods} 
The performances obtained by OPAL are compared to 
other methods applied to the same dataset in Tables~\ref{tab:compICBM} and~\ref{tab:compADNI}.
The presented values are the results
published by the authors. 
The provided computation times are the times dedicated to segmentation step only but
do not include template preselection while only OPAL does not require it.
Therefore,
the computation times are under-estimated except for
OPAL.

On the ICBM dataset, compared to the original PBL \cite{coupe2011patch}, 
OPAL improves segmentation accuracy by 1.7 percentage points (pp) while being $700\times$ faster.  
Compared to the most accurate method on this dataset, based on dictionary learning (DDLS \cite{tong2013segmentation}), 
OPAL obtained higher Dice coefficients
for computation times $1000\times$ faster and
with a $p$-value inferior to $10^{-12}$ obtained from a paired t-test on the OPAL and DDLS sets of Dice coefficients.
In addition, for a given Dice coefficient of $89.0\%$ (equivalent
to the DDLS method accuracy)  OPAL requires less than $0.22$s on the ICBM dataset ($4000\times$ faster than DDLS method). 

On the EADC-ADNI dataset, OPAL results are compared to other methods only in terms of
segmentation accuracy, since computation times are not provided by the authors in their publications.
The results presented with OPAL on EADC-ADNI in Table~\ref{tab:compADNI}
are obtained in $1.51$s processing per subject. 
In all studied cases, OPAL produced the best segmentation accuracy with a mean Dice coefficient of $89.8\%$ (median Dice of $90.1\%$). 
The Dice values show that OPAL outperforms recently proposed methods on EADC-ADNI. 
Indeed, compared to a Random forest approach \cite{tangaro2014}, OPAL improves segmentation accuracy by 13.8pp and compared to recent multi-template approaches OPAL obtained a gain superior to 2.2pp, 
with a $p$-value inferior to $10^{-25}$ obtained from a paired t-test on the OPAL and LEAP sets of Dice coefficients.

\begin{table}[h!]
\caption{Method comparison in terms of segmentation accuracy and
  computation time (per subject) for the ICBM dataset.}
\label{tab:compICBM}
\centering
\newcommand{\sz}{\hspace{8pt}}
\newcommand{\szz}{\hspace{5pt}}
\newcommand{\szzz}{\hspace{15pt}}
{\small
\begin{tabular}{@{\szz}l@{\szzz}c@{\sz}c@{\sz}c@{\sz}r@{\szz}}
\hline
\multirow{1}{*}{\small{Method on ICBM}}&\multirow{1}{*}{Median Dice}&\multirow{1}{*}{$95\%$ interval}&\multicolumn{1}{c}{Comp. Time}\\
\hline
Patch-based (PBL)\cite{coupe2011patch} & $88.2\pm2.19\%$ &$[87.7;88.7]\%$& $662s$  $({\times}700)$\\
Multi-templates (MTM)\cite{collins2010towards}   & $88.6\pm2.05\%$ &$[88.2;89.0]\%$& $3974s$ $({\times}4300)$\\
Sparse coding (SRC)\cite{tong2013segmentation} & $88.7\pm1.94\%$  & $[88.3;89.2]\%$ & $5587s$ $({\times}6000)$\\
Dictionary learning (DDLS)\cite{tong2013segmentation}& $89.0\pm1.90\%$ & $[88.5;89.4]\%$ & $943s$  $({\times}1000)$\\
\textbf{OPAL}& $\mathbf{89.9\pm1.70\%}$ &$\mathbf{[89.6;90.3]\%}$&  $\mathbf{0.92s}$\\
\hline
\end{tabular}
}
\end{table}

\begin{table}[h!]
\caption{Method comparison in terms of segmentation accuracy for the EADC-ADNI dataset. 
  Since none of the selected publications mention their computation times, the
  comparison only focus on the mean Dice coefficient. The selected result for OPAL method
  was obtained in $1.51$s processing per subject.}
\label{tab:compADNI}
\centering
\newcommand{\sz}{\hspace{8pt}}
\newcommand{\szz}{\hspace{4pt}}
\newcommand{\szzz}{\hspace{15pt}}
{\small
\begin{tabular}{@{\szz}l@{\szzz}c@{\sz}c@{\sz}c@{\szz}}
\hline
\multirow{1}{*}{\small{Method on EADC-ADNI}}&\multirow{1}{*}{Mean Dice}&\multirow{1}{*}{$95\%$ interval}\\
\hline
Random Forest \cite{tangaro2014} & $76.0\pm7.00\%$ &$[74.6;77.4]\%$\\
Multi-templates (BMAS)\cite{roche2014}   & $86.6\pm1.70\%$ &$[86.3;86.9]\%$\\
Multi-templates (LEAP)\cite{gray2014}  & $87.6\pm2.07\%$ &$[87.1;88.0]\%$\\
\textbf{OPAL}& $\mathbf{89.8\pm1.46\%}$  & $\mathbf{[89.5;90.1]\%}$\\
\hline
\end{tabular}
}
\end{table}

\subsection{Complementary Results}

\noindent\textbf{Automatic segmentations as priors.}
Recently, several works have proposed to use automatic segmentations as priors in 
order to accurately segment a new subject. A way to improve segmentation accuracy 
consists in increasing the size of the template library.
In order to do this, subjects without expert segmentations 
are automatically segmented and added to the
template library of manually segmented subjects \cite{eskildsen2012beast}.
The Multiple Automatically Generated Templates (MAGeT) approach has been proposed in \cite{pipitone2014}
and works by propagating segmentations to a template library, 
composed of a subset of unlabeled subjects, 
via transformations estimated by nonlinear registrations. 
The resulting segmentations are then used as template library to segment a new subject.
Similarly, the LEAP method \cite{gray2014} proposes to
propagate the label segmentation to unlabeled subjects by iteratively segmenting the 
closest subjects in terms of joint entropy. 
These approaches lead to segmentation accuracy improvement, since the diversity of the dataset used to segment a subject is increased.

As mentioned in section~\ref{subsubsection:OPM}, the computation time and 
complexity of OPAL only depends on the size of the subject to segment.
This important fact enables us to extend the library size with no impact on the complexity of the algorithm.
New subjects without manual expert segmentations can be automatically segmented and
 added to the template library in order to improve its diversity.
Consequently, the segmentation accuracy of a new subject may be improved, since more relevant
matches can be found within the template library.

\begin{figure}[H]
\centering
\newcommand{\siz}{0.85\textwidth}
\fbox{\includegraphics[width=\siz, height=120pt]{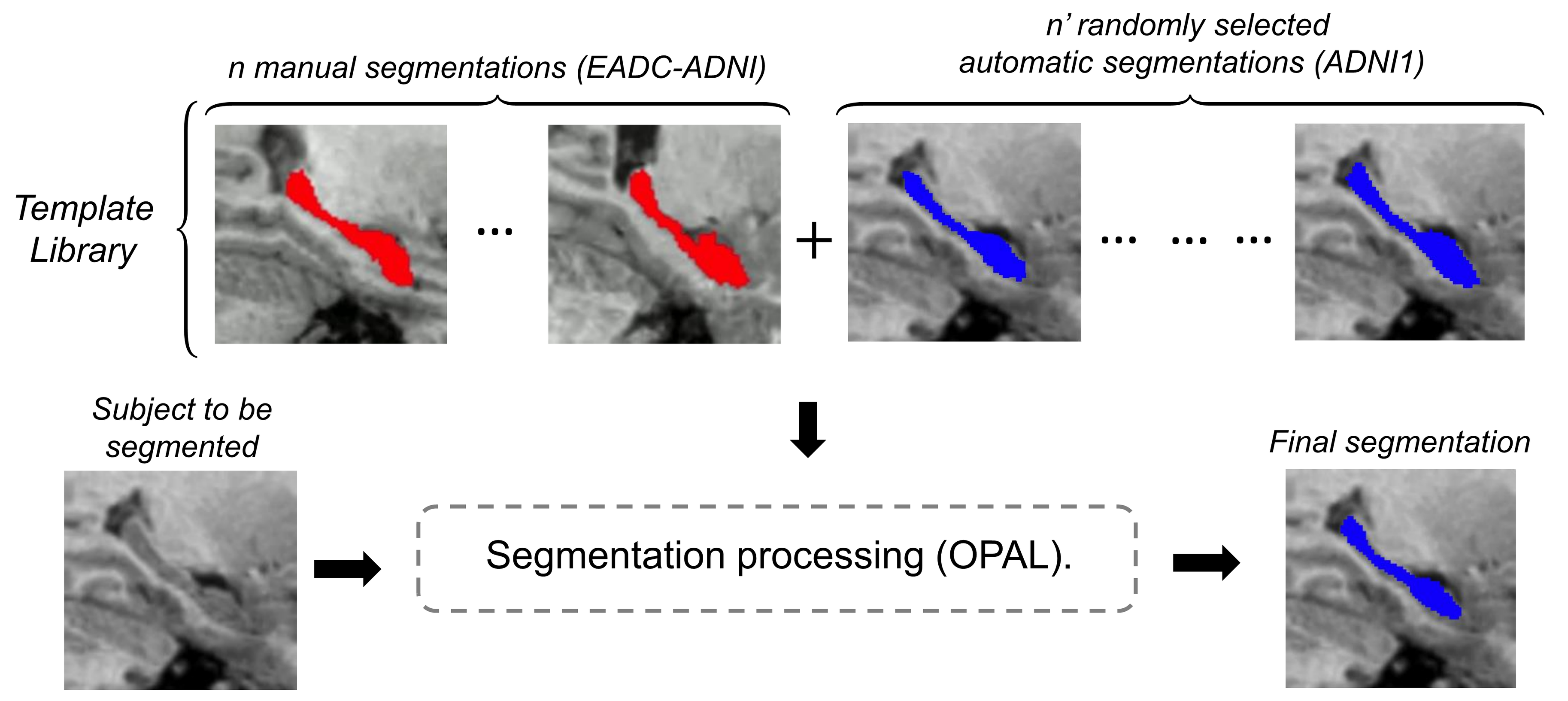}}
\caption{Addition of new segmented subjects to the template library.
The automatic segmentation of new subjects provided without manual expert segmentations
can be added to the template library in order to increase its size and diversity.
Consequently, later segmentations may benefit from more numerous and potentially better training templates.
} 
\label{fig:addsubjects}
\end{figure}

\begin{figure}[H]
\centering
\newcommand{\siz}{0.47\textwidth}
\newcommand{\sizz}{120pt}
\includegraphics[width=\siz, height=\sizz]{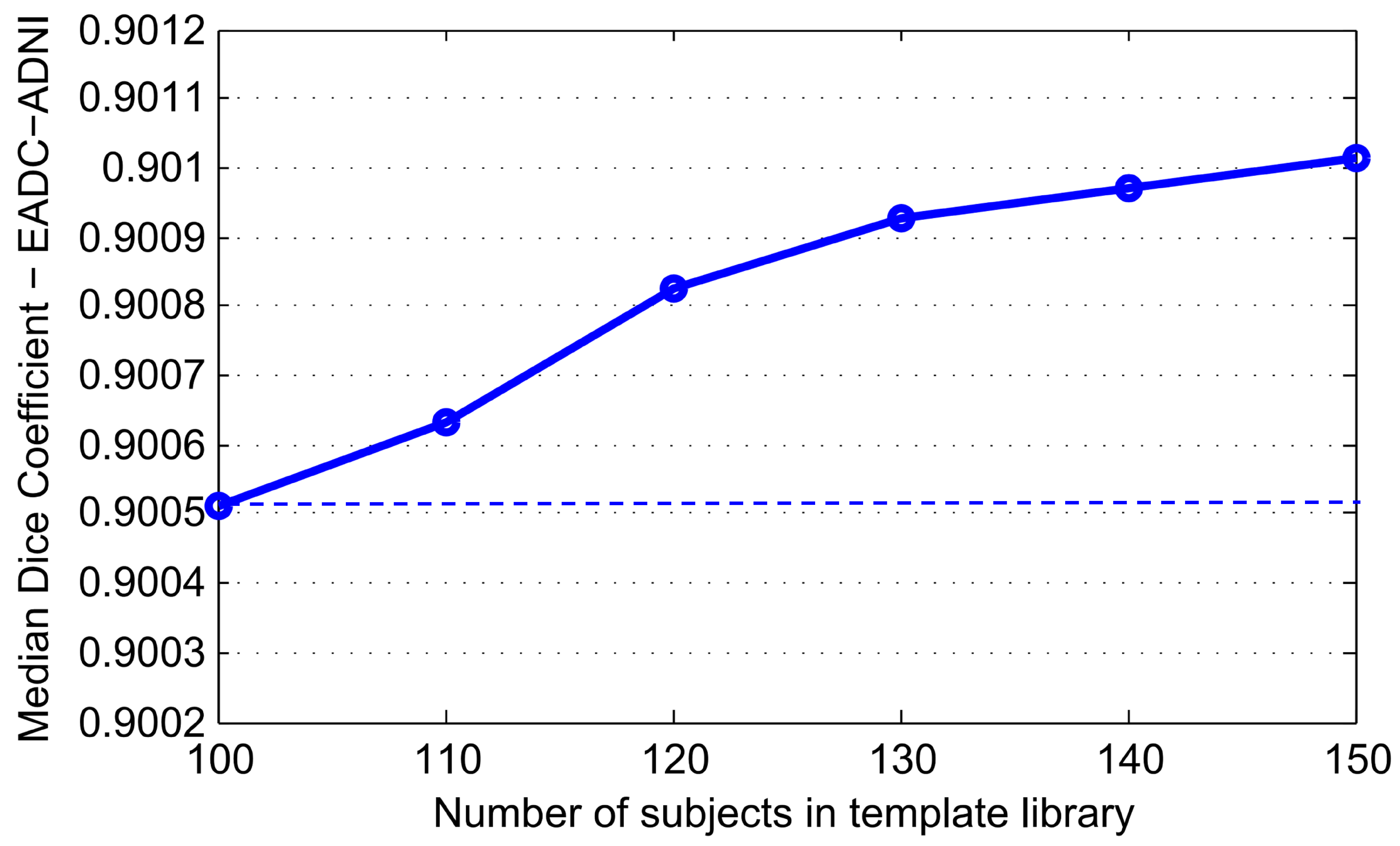}
\includegraphics[width=\siz, height=122pt]{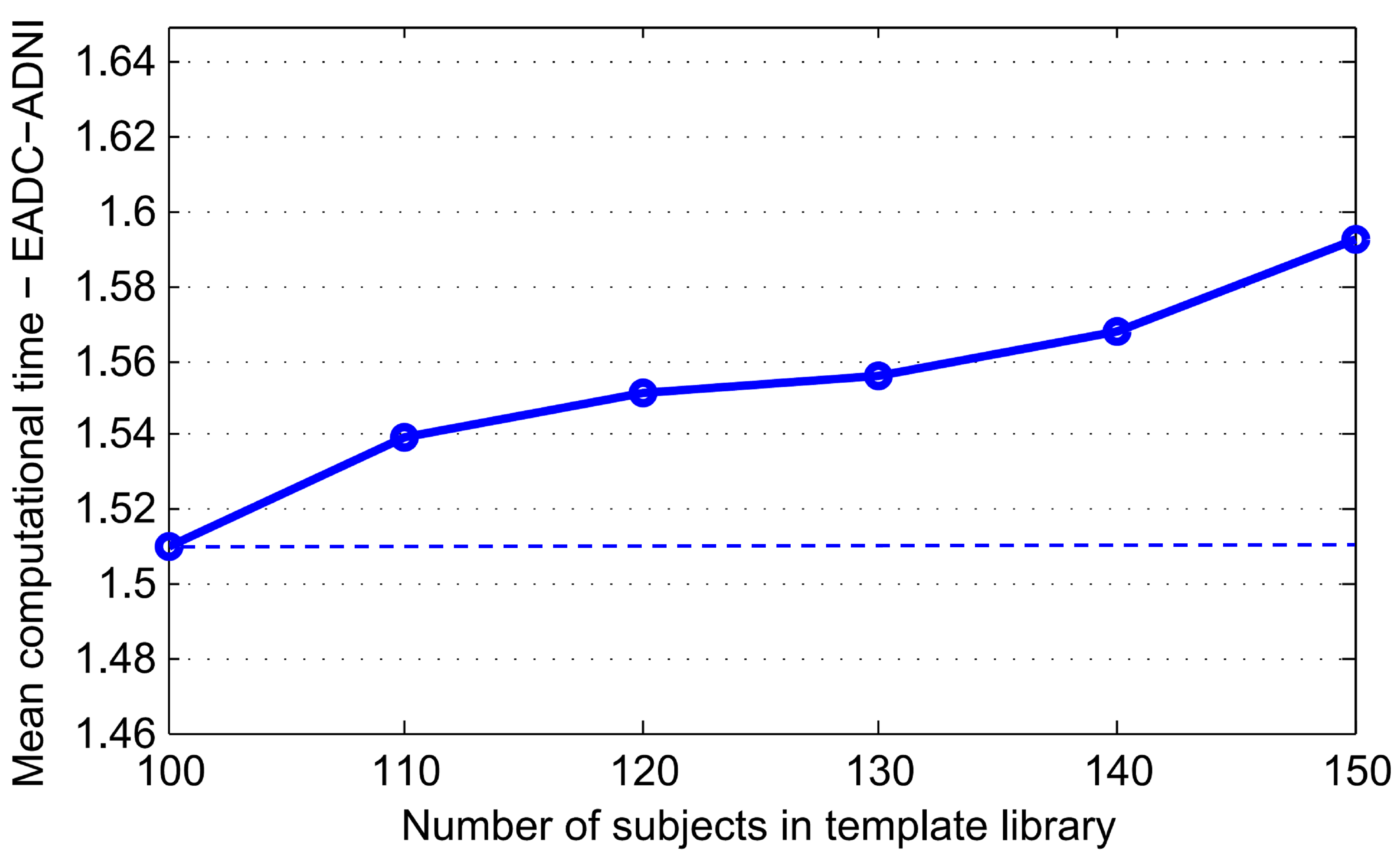}
\caption{
Influence of the addition of automatic segmented ADNI subjects 
to the EADC-ADNI dataset on the segmentation accuracy 
(left) and the corresponding computation time (right). 
The results obtained with $100$ subjects (dotted line) correspond to the selected
results in Table~\ref{table:contribADNI}.
} 
\label{fig:autorefaddS}
\end{figure}

Therefore, we propose an experiment where automatically segmented subjects from the standardized ADNI1 dataset \cite{wyman2013standardization} 
are randomly selected and added to the EADC-ADNI template library as illustrated in Figure~\ref{fig:addsubjects}.
The Dice coefficient is still computed with a leave-one-out procedure on
the EADC-ADNI subjects with provided expert-based segmentations.
Figure~\ref{fig:autorefaddS} shows the impact of 
increasing the library size,
on the segmentation accuracy and computation time.

Adding new templates to the library with automatic segmentations as priors
enables us to improve the segmentation accuracy. 
Indeed, since the dataset is extended with new subjects,
its diversity is increased and more relevant
matches can be found within the template library.
Most importantly, the computation time results in Figure~\ref{fig:addsubjects} 
highlight the important fact 
that  OPAL complexity only depends
on the size of the subject to segment and not on the size of the template library.
Adding subjects to the database 
improves the segmentation accuracy at the expense of a very little setback on 
computation time (due to memory storage and data transfer).
With $50\%$ of supplementary training templates, the computation time is only increased by $6\%$.

\begin{figure}[H]
\centering
\newcommand{\siz}{0.48\textwidth}
\newcommand{\sizz}{130pt}
\newcommand{\sizzz}{0.48\textwidth}
\subfigure{\includegraphics[width=\siz, height=\sizz]{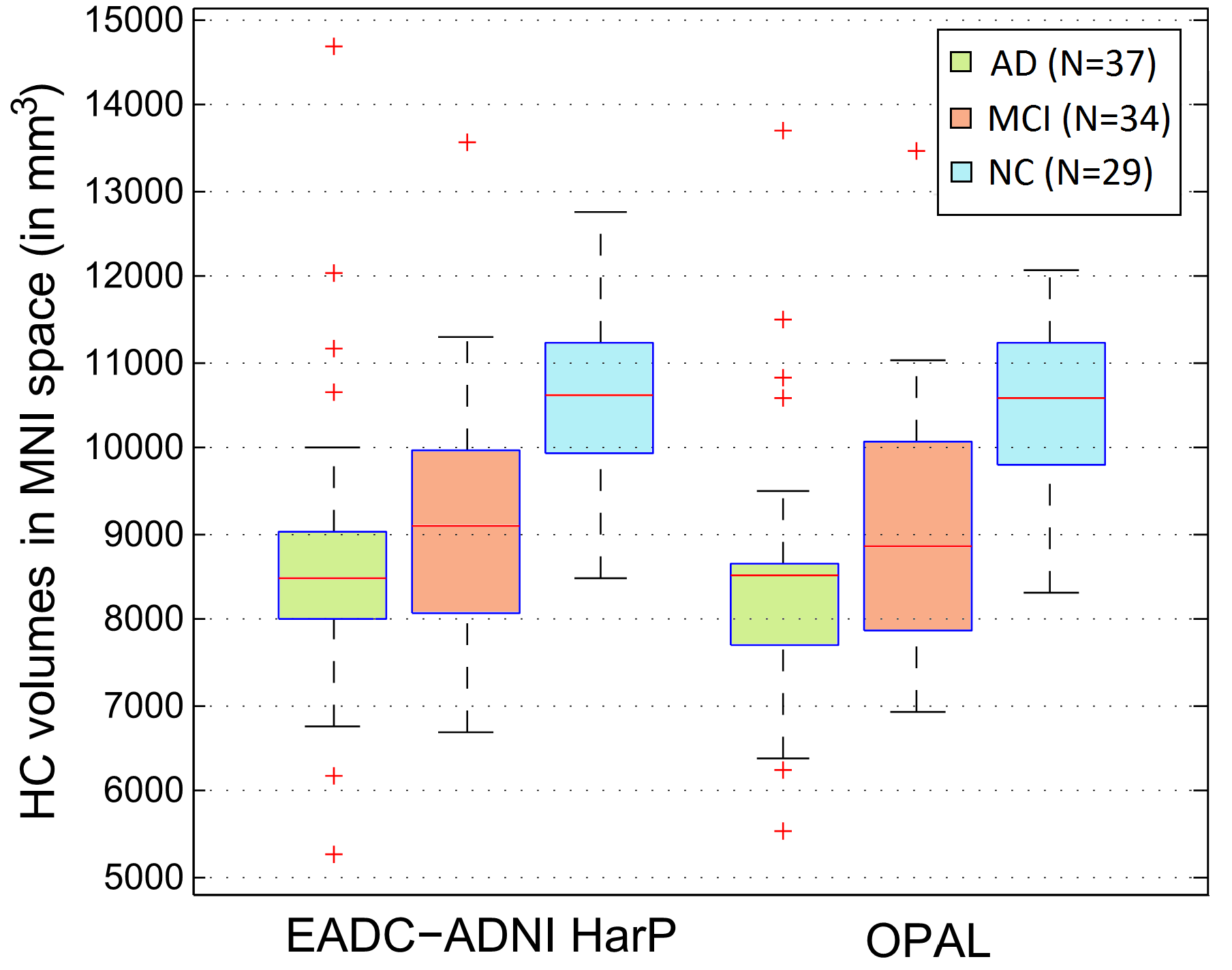}}
\subfigure{\includegraphics[width=\sizzz, height=135pt]{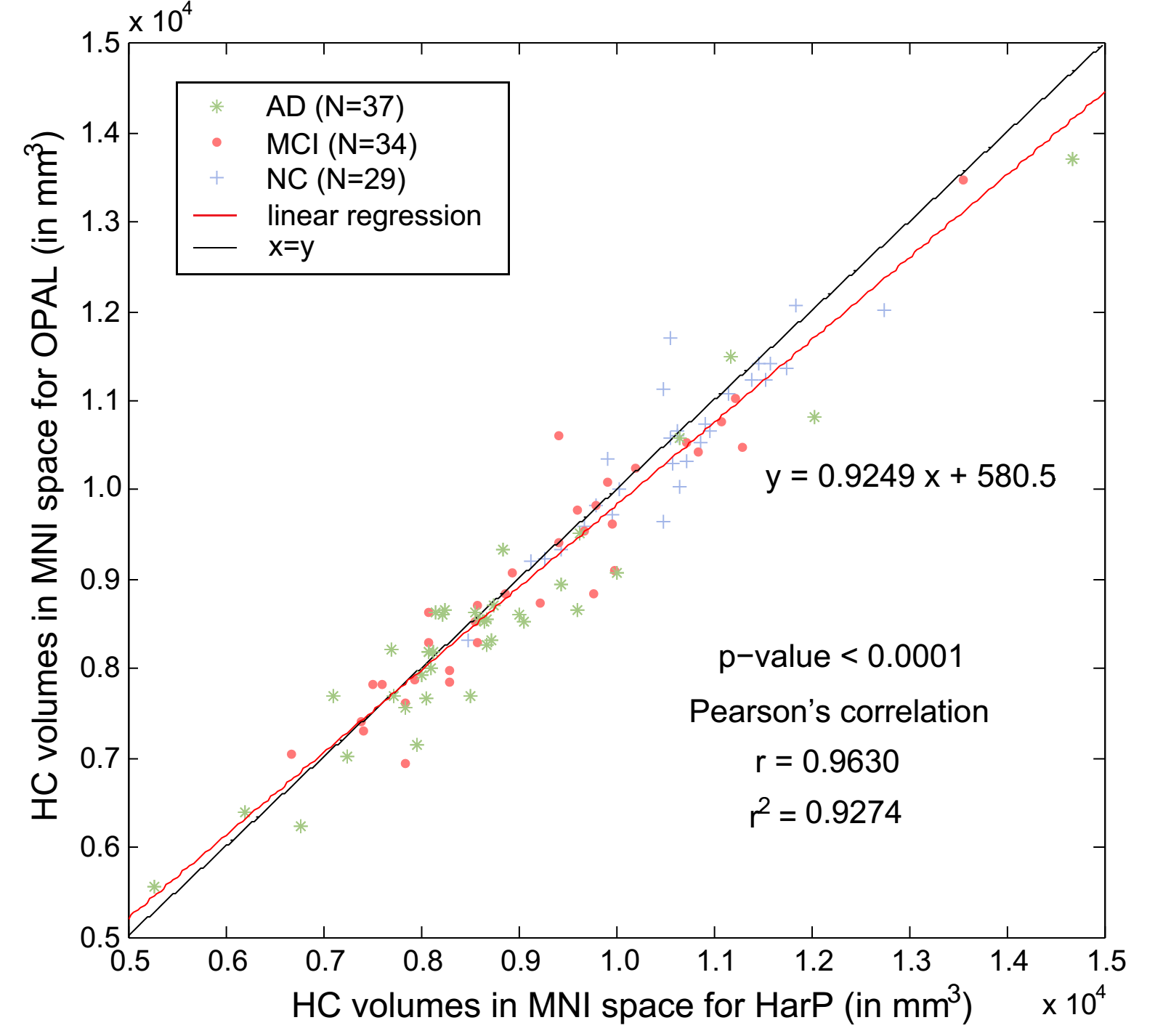}}
\caption{Hippocampal volumes in the MNI space of the segmentation results from reference EADC-ADNI harmonized protocol and OPAL method (left). 
Correlation between hippocampal volumes of HarP and OPAL segmentations (right).
} 
\label{fig:boxplot}
\end{figure}

\noindent\textbf{Clinical application.}
Finally, we propose to show the performance of our method on a clinical application,
by comparing population separation accuracy using manual segmentation of the EADC-ADNI harmonized protocol (HarP) \cite{boccardi2014delphi} 
and the OPAL segmentation.
The area under the ROC curve (AUC) is computed on
hippocampal volumes in the MNI space for both manual and OPAL segmentation results 
on the three groups of the EADC-ADNI dataset, 
AD (Alzheimer's Disease, N=37), MCI (Mild Cognitive Impairment, N=34) and NC (Normal Controls, N=29).
As shown in Table~\ref{table:AUC_ADNI}, the segmentation results provided by OPAL enable to better separate groups with a higher AUC.
The Pearson's correlation is also computed between the HarP and OPAL hippocampal volumes of segmentations. 
In Figure \ref{fig:boxplot}, the hippocampal volumes distribution for each group are represented.
The correlation between hippocampal volumes of HarP and OPAL segmentations is also illustrated.

\begin{table}[h!]
\caption{
Area under the ROC curve (AUC) on hippocampal volumes in the MNI space of the segmentation results from reference EADC-ADNI harmonized protocol 
and OPAL method.
}
\label{table:AUC_ADNI}
\centering
\newcommand{\sz}{\hspace{10pt}}
\newcommand{\szz}{\hspace{5pt}}
\newcommand{\szzz}{\hspace{15pt}}
{\small
\begin{tabular}{@{\szz}l@{\szzz}c@{\sz}c@{\szz}}
\hline
\multirow{1}{*}{} &\multirow{1}{*}{EADC-ADNI HarP}&\multirow{1}{*}{OPAL} \\
\hline
HC mean
volume (mm\textsuperscript{3}) &$9397\pm1588$&$9272\pm1525$\\
AUC NC vs. AD &$0.884$&$0.898$\\
AUC NC vs. MCI  &$0.805$&$0.821$\\
AUC MCI vs. AD  &$0.612$&$0.634$\\
\hline
\end{tabular}
}
\end{table}

\section{Discussion}

Our proposed OPAL method presents several differences with
state-of-the-art PBL approaches. 
First, the complexity of the optimized PatchMatch algorithm (see Figure~\ref{fig:opal}) only depends on the size of subject's image.
Consequently, the entire image library
$T$ is used without any template preselection step,
at constant complexity in time.
The linear registration is also exploited by constraining the search for patch matches at each step.
Secondly, a patchwise label fusion is performed from the selected matches (see Figure~\ref{fig:method1}) and a bilateral kernel is also
used to increase spatial consistency leading to better segmentation results, as done in \cite{manjon2014}.
Finally, we introduced a new multi-scale and multi-feature framework based on late aggregation of estimators.
This new approach is possible thanks to the very low computational burden of the ANN search in our OPM framework.
Independent multi-scale and multi-feature ANN searches are carried out, and 
a late fusion is finally performed on all resulting estimator maps from PBL to produce
the final segmentation as illustrated in Figure~\ref{fig:method2}.
We validated our method on two datasets for hippocampus segmentation. 
These datasets cover different manual segmentation protocols and preprocessing pipeline. 
By this way, the robustness of OPAL to hippocampus definition and processing has been studied.

On ICBM and EADC-ADNI datasets, we respectively obtained a median Dice coefficient
of $89.9\%$ and $90.1\%$ in approximately $1.5$s processing per subject. 
A large comparison with published methods such as original
PBL \cite{coupe2011patch}, 
sparse representation (SRC) \cite{tong2013segmentation}, 
dictionary learning (DDLS)
\cite{tong2013segmentation}, 
multi-templates (MTM, BMAS, LEAP) \cite{collins2010towards, roche2014, gray2014}
and random forest \cite{tangaro2014},
highlights the very 
competitive results of the proposed method (see Tables~\ref{tab:compICBM} and~\ref{tab:compADNI}).

For the EADC-ADNI comparison, the computation times are not provided by the authors.
However, we may assume that the BMAS~\cite{roche2014} and LEAP~\cite{gray2014} methods are likely to propose comparable computation time to MTM~\cite{collins2010towards} since they are also
based on a multi-templates warping approach. One can note that multi-templates warping methods perform worse on the EADC-ADNI dataset than on the ICBM dataset. This can be related to higher anatomical variability in EADC-ADNI dataset due to the presence of Alzheimer's disease (AD). On this dataset, the well defined one-to-many mapping offering by patch-based segmentation appears to better capture this higher variability. 

It is important to note OPAL can reach the inter-expert
reliability on both datasets ($90\%$ and $89.0\%$ respectively for ICBM and EADC-ADNI datasets).
Moreover, this has been validated on two datasets with two different manual segmentation protocols. 
While more than 30 minutes are required by an expert to segment one hippocampus (1 hour for both), 
OPAL produces similar segmentation quality in less than $2$s.
OPAL is performed on denoised and registered images that are preprocessed in less than $5$min (see section~\ref{subsection:dataset}).
We compared the population separation accuracy using manual segmentation of HarP protocol and OPAL segmentation. 
The robustness and consistency of our automatic segmentation method enable a better group separation between ADNI
populations (AD, MCI, NC).
Complementary results on the use of automatic segmentations as priors have been also presented. 
We show that improvements can be obtained without significant increasing of computation time by adding subjects to the training library.

Throughout this paper, we mentioned OPAL high capacities in terms of both segmentation 
and computation time. With such fast performance, OPAL opens the way for new
applications of label fusion segmentation such as integration in visualization 
software that would highly facilitate the analysis of brain MRI.
A web-based tool for on-line remote MRI processing is also 
a possible application to exploit OPAL capacities. We plan to include OPAL in the next version of volBrain (\url{http://volbrain.upv.es}).

Finally, in this paper we only applied our method to the hippocampus segmentation, 
since it is the most studied structure in the Alzheimer's disease context.
Nevertheless, the OPAL method can be applied to the segmentation of any anatomical structure.
Future research will focus on the extension of the method to the whole brain segmentation as done in 
\cite{heckemann2006automatic}. Our preliminary results suggest that this can be done in less than 2 minutes.

%

\section{Conclusion}

In this paper, we propose a novel patch-based segmentation method
based on an optimized PatchMatch label fusion.
Thanks to the low computational burden of our method, we investigated the potential of a new
multi-feature and multi-scale framework with late estimator aggregation.
The validation of our approach 
on hippocampus segmentation applied to two different datasets
shows that the proposed method
produces competitive results compared to the state-of-the-art approaches. Indeed,
OPAL obtained the highest median Dice coefficient with a drastically reduced computation time. In addition, OPAL reaches the inter-expert
reliability on both datasets ($90\%$ and $89.0\%$ respectively for ICBM and EADC-ADNI datasets).
Therefore, OPAL  provides automatic segmentations
equivalent in terms of Dice coefficient to inter-expert segmentations 
in less than $2$s of processing for the segmentation step.
In addition, the volumes segmented by OPAL are highly correlated to the manually segmented volumes. 
Finally, the accuracy and reproducibility of OPAL enable to better separate ADNI groups (AD, MCI, NC).


\section*{Acknowledgments}

This study has been carried out with financial support from the French 
State, managed by the French National Research Agency (ANR) in the  
frame of the Investments for the future Program IdEx Bordeaux 
(ANR-10-IDEX-03-02), Cluster of excellence CPU and TRAIL (HR-DTI
ANR-10-LABX-57). 
We also thank Tong Tong and Daniel Rueckert 
for providing us complete results of the methods proposed in \cite{tong2013segmentation}, 
Sonia Tangaro 
and Marina Boccardi 
for providing us complete results of the method proposed in \cite{tangaro2014}, 
and
Katherine Gray and Robin Wolz
for providing us complete results of the LEAP method proposed in \cite{gray2014}.
Data collection and sharing for this project were funded by the Alzheimer's Disease Neuroimaging Initiative (ADNI) 
(National Institutes of Health Grant U01 AG024904). The ADNI is funded by the National Institute on Aging and the National 
Institute of Biomedical Imaging and Bioengineering and through generous contributions from the following: Abbott, AstraZeneca 
AB, Bayer Schering Pharma AG, Bristol-Myers Squibb, Eisai Global Clinical Development, Elan Corporation, Genentech, 
GE Healthcare, GlaxoSmithKline, Innogenetics NV, Johnson and Johnson, Eli Lilly and Co., Medpace, Inc., Merck and Co., Inc., 
Novartis AG, Pfizer Inc., F. Hoffmann-La Roche, Schering-Plough, Synarc Inc., as well as nonprofit partners, 
the Alzheimer's Association and Alzheimer's Drug Discovery Foundation, with participation from the U.S. Food and Drug Administration. 
Private sector contributions to the ADNI are facilitated by the Foundation for the National Institutes of Health (www.fnih.org). 
The grantee organization is the Northern California Institute for Research and Education, and the study was coordinated by the Alzheimer's Disease 
Cooperative Study at the University of California, San Diego. ADNI data are disseminated by the Laboratory for Neuro Imaging at the University of California,
Los Angeles.
This research was also supported by the Spanish grant TIN2013-43457-R from the Ministerio de Economia y competitividad, 
NIH grants P30AG010129, K01 AG030514 and the Dana Foundation.

\section*{References}

\bibliographystyle{unsrt}
\bibliography{NIMG-15-937}

\end{document}